\documentclass[runningheads]{llncs}

\usepackage{eccv}

\usepackage{eccvabbrv}

\usepackage[table,xcdraw]{xcolor}
\newcommand{\minisection}[1]{\vspace{0.5mm} \noindent \textbf{#1} \hspace{0mm}}
\usepackage{arydshln} 
\usepackage{tikz}
\usetikzlibrary{positioning}
\newcommand{\vect}[1]{\boldsymbol{#1}}
\usepackage{adjustbox}     
\usepackage{multirow}
\definecolor{mygray}{gray}{0.95}
\usepackage{hhline}
\usepackage{nicematrix}
\usepackage{amssymb}
\usepackage{pifont}

\usepackage{tabularx}
\usepackage{booktabs}
\definecolor{mygray}{gray}{0.87}
\usepackage{booktabs} 

\usepackage{graphicx}
\usepackage{booktabs}

\usepackage{orcidlink}
\definecolor{cvprblue}{rgb}{0.21,0.49,0.74}

\begin{document}

\title{FlowMark: Mask-Guided Video Watermarking}

\titlerunning{FlowMark}

\author{Vishal Asnani$^{1}$ \and
Shruti Agarwal$^{1}$ \and
John Collomosse$^{1,2}$}
\authorrunning{ Asnani et al.}
\institute{$^{1}$Adobe Research, $^{2}$DECaDE, University of Surrey \\
\email{\{vasnani, shragarw, collomos\}@adobe.com}}

\maketitle

\sloppy
\begin{abstract}
We present FlowMark, a video watermarking framework guided by automatically predicted object masks. In contrast to prior region-based approaches that require user-supplied mask guidance, FlowMark learns to identify optimal regions for watermark embedding through a dedicated Mask Predictor network. Our end-to-end trainable architecture combines region-aware encoding with noise-augmented training to ensure robustness against compression, geometric transformations, and content variation, while preserving high perceptual quality. Our content-adaptive masking keeps watermark signals coherent with natural video dynamics, effectively eliminating perceptual flicker. Beyond compression robustness, FlowMark maintains reliable watermark recovery under video-native temporal edits (e.g., frame swap, insertion, deletion, resampling, and interpolation) and real-world social media distribution pipelines (e.g., YouTube and Facebook re-encoding). Experimental results on both image and video datasets show that FlowMark reliably embeds $128$-bit messages with up to $50.08$ dB PSNR, offering strong performance for content provenance, temporal authenticity verification, and video integrity protection.
\end{abstract}    
\section{Introduction}
\label{sec:intro}

Advances in generative video are enabling highly realistic synthetic content, unlocking powerful creative tools but also raising concerns around misuse. A common mitigation is to attach provenance metadata that raises awareness of content origins (e.g., Sora and Veo3 attach C2PA metadata~\cite{c2pa}). For images, such metadata is often reinforced through invisible watermarking~\cite{Collomosse2024,synthid}, embedding durable identifiers that persist through format conversion and content platforms.

\begin{figure}[t]
    \centering
    \includegraphics[width=1\columnwidth]{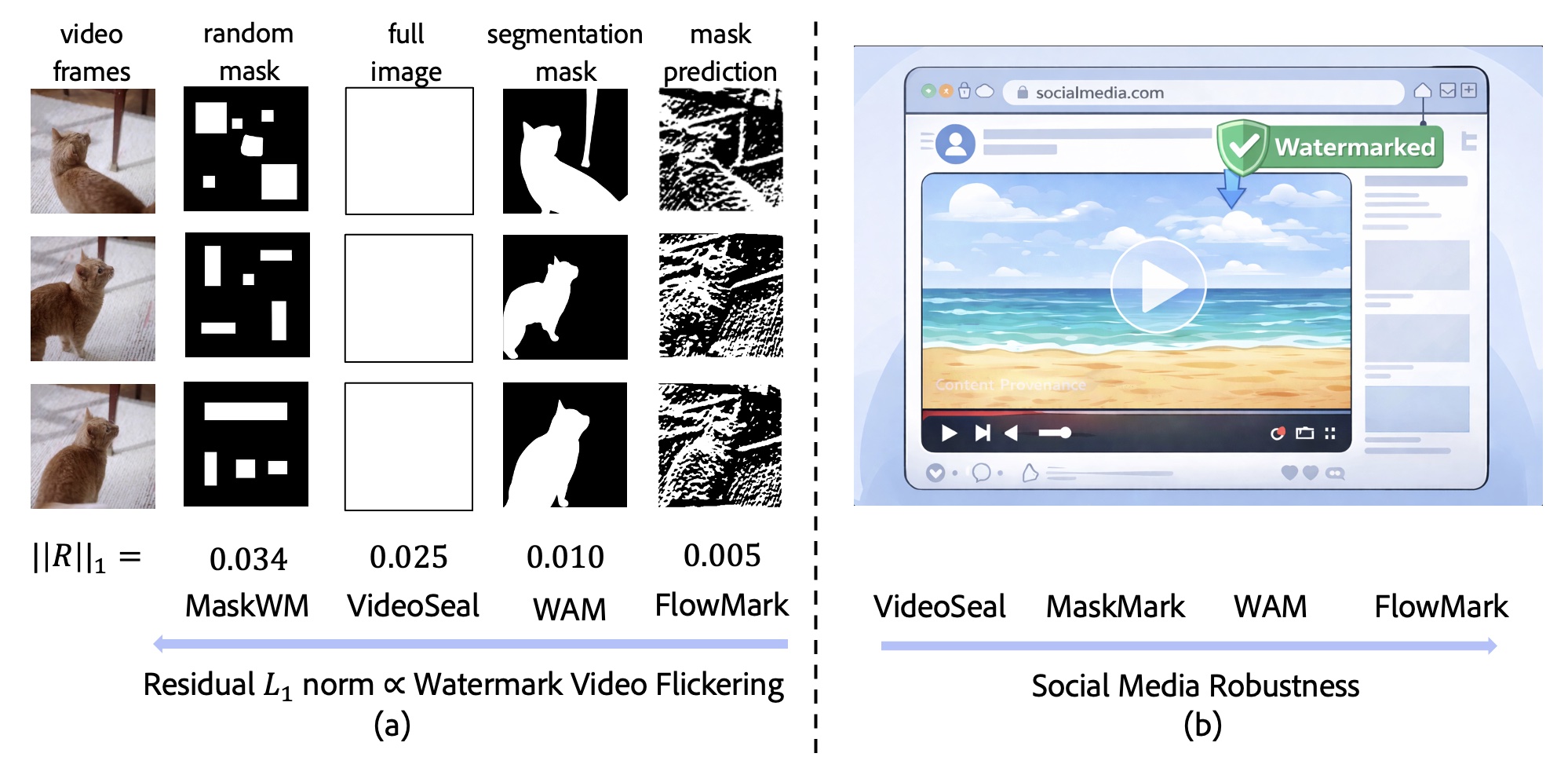}
    \caption{
        (a) VideoSeal~\cite{fernandez2024video} embeds watermarks across the full frame, 
        WAM~\cite{sander2025wam} uses segmentation masks, and MaskWM~\cite{hu2025maskimagewatermarking} relies on random masks. 
        FlowMark learns content-aware masks that place the watermark in perceptually resilient regions. 
        The residual \(R\) is the pixel-wise difference between the original and watermarked frames, and a lower \(\lVert R \rVert_{1}\) indicates less visible change and reduced flickering across consecutive video frames. 
        FlowMark achieves the lowest residual norm among all methods. (b) FlowMark achieves highest watermark detection robustness on social media websites.
    }
    \label{fig:mask_comparison}
    \vspace{-6mm}
\end{figure}

Extending watermarking to video presents unique challenges beyond those encountered in image watermarking. Most existing approaches rely on global spatial or frequency-domain embedding~\cite{zhu2018hidden, bui2025trustmark}, and when applied independently to video frames, they often introduce temporal inconsistencies such as flicker due to a lack of coherence across frames. These artifacts become especially pronounced under the spatio-temporal distortions introduced by modern video codecs like H.264 and H.265, resulting in fragile watermark recovery. While recent techniques such as VideoSeal~\cite{VideoSeal2024} are trained for robustness against compression, they still suffer from visible flickering because the embedded watermark can vary from frame to frame. Other approaches like WAM~\cite{sander2025wam} and MaskWM~\cite{hu2025mask} incorporate Just Noticeable Difference (JND) models~\cite{wu2017enhanced} and per-frame region masking to improve perceptual quality, but these require manual mask selection during encoding, a process that is impractical for real-world video pipelines. These approaches embed signals globally or within fixed segmentation or random masks, often leading to uneven watermark energy and higher residual $L_1$ norms across frames (~\cref{fig:mask_comparison}).
A related strategy for mitigating flicker in neural watermarking is to embed the \textit{same} watermark message across multiple consecutive frames~\cite{fernandez2024video}. Although this stabilizes appearance, it inherently restricts frame-level uniqueness, preventing fine-grained temporal provenance. For authentic media attribution, each frame should retain its own verifiable signature while preserving temporal coherence. 

In this work, we ask: Can a model learn the optimal spatial mask for watermark embedding automatically, frame by frame? We answer this by introducing FlowMark, a mask-guided video watermarking framework designed for temporally stable, perceptually adaptive, and compression-robust embedding. FlowMark learns spatial embedding regions dynamically through a lightweight mask predictor, enabling content-aware watermark placement. To enhance perceptual quality, we introduce dark-region adaptive residual modulation, which attenuates watermark strength in low-luminance areas where compression artifacts are most noticeable. Temporal stability is enforced via a temporal change consistency objective that preserves frame-to-frame motion dynamics and reduces flicker.

To ensure robustness under real-world distribution pipelines, FlowMark is trained using a three-stage curriculum that progressively transitions from clean pretraining to image-level augmentations and finally to video codec distortions (H.264, H.265, VP9, AV1), simulating social media compression (YouTube and Facebook) and post-processing. In addition to compression robustness, FlowMark is resilient to video-specific temporal operations commonly encountered in editing pipelines, including frame reordering, insertion, deletion, resampling, and interpolation. We further adopt a structured bit-embedding scheme encoding both a persistent video identifier and a frame index, enabling frame-level traceability under such temporal edits. Extensive experiments demonstrate strong robustness, perceptual quality, and temporal smoothness across image and video benchmarks as well as temporal-edit scenarios.

\begin{enumerate}
    \item We propose FlowMark, a mask-guided video watermarking framework that automatically learns spatially adaptive watermark regions for temporally stable watermark insertion.
    \item We introduce dark-region adaptive residual modulation and temporal change consistency constraints to jointly improve perceptual quality and motion stability.
    \item We design a three-stage curriculum training strategy that progressively incorporates image-level and codec-level distortions for robust real-world deployment of video watermarking on social media.
    \item We develop a structured per-frame bit embedding scheme that encodes both video identity and frame indices, enabling reliable detection of video temporal edits.
\end{enumerate}

\section{Related Works}

\minisection{Image Watermarking.}
Image watermarking has been extensively explored for tasks such as copyright protection~\cite{wan2022comprehensive}, manipulation or AI-content detection~\cite{zhang2025omniguard,asnani2022proactive}, attribution~\cite{asnani2024promark,asnani2025custommark}, and provenance verification~\cite{bui2025trustmark,videosealarxiv}. 
Early methods modified pixel intensities using least significant bit (LSB) substitution~\cite{wolfgang1996watermark}, or embedded information in transform domains such as DCT, DWT, or hybrid DWT–DCT–SVD~\cite{taha2022high,ghazanfari2011lsb++,navas2008dwt}.  
Modern GenAI-oriented systems like SynthID~\cite{synthid}, Stable Signature~\cite{fernandez2023stable}, and TreeRing~\cite{wen2023treerings} focus on binary content detection rather than message decoding.  
The rise of deep learning watermarking began with HiDDeN~\cite{zhu2018hidden}, which introduced a CNN encoder–decoder trained with differentiable noise layers.  
Subsequent methods improved robustness and imperceptibility via geometric regularization (StegaStamp~\cite{tancik2020stegastamp}), latent-space embedding (SSL~\cite{fernandez2022watermarking}, RoSteALS~\cite{bui2023rosteals}), and perceptual loss optimization (TrustMark~\cite{trustmarkarxiv,bui2025trustmark}, InvisMark~\cite{xu2025invismark}).  
MBRS~\cite{mbrs} and CIN~\cite{ma2022towards} further enhance compression robustness by simulating codec-induced noise during training, while recent region-based methods such as WAM~\cite{sander2025wam} and Mask Image Watermarking~\cite{hu2025maskimagewatermarking} enable localized watermarking using predicted segmentation masks, though their payload capacity and spatial consistency remain limited.

\minisection{Video Watermarking.}
Classical video watermarking methods can be grouped into \textit{codec-integrated} and \textit{spatial-domain} approaches.  
Codec-integrated techniques embed watermarks into compressed streams such as MPEG-2, H.264, or H.265 by modifying entropy codes or motion vectors~\cite{Hartung1998,Yong2008,Alattar2003,Zhang2007,Guo2010,Tew2016,Dutta2018}.  
While efficient and visually imperceptible, these methods are fragile under transcoding and tightly coupled to specific codecs.  
Spatial-domain approaches instead embed signals in decoded frames via pixel or frequency transformations~\cite{Coria2008}, achieving codec independence but limited robustness to compression and geometric distortions.

Deep-learning-based methods address these issues by learning end-to-end encoders and decoders that simulate distortions during training.  
VStegNet~\cite{mishra2019vstegnet} and RivaGAN~\cite{RivaGAN2019} extended image watermarking to video but suffered from scalability constraints.  
DVMark~\cite{DVMark2023}, VHNet~\cite{shen2023vhnet}, and REVMark~\cite{zhang2023novel} introduced differentiable compression modules for robustness, while RIVIE~\cite{jia2022rivie} and V2A-Mark~\cite{zhang2024v2a} incorporated temporal losses and dual-watermark designs.  
ItoV~\cite{ye2023itov} reformulated 2D CNNs to process videos by merging temporal and channel dimensions, improving efficiency but limiting sequence length.  
VideoSeal~\cite{fernandez2024video} advanced this direction through differentiable distortion pipelines that enhance general robustness.  

Despite these advances, existing methods either apply watermark residual globally or depend on codec-specific operations, often producing visible flicker and spatial artifacts.  
\textit{FlowMark} introduces a mask-guided embedding strategy that adaptively localizes watermark regions aligned with motion and texture, achieving codec-agnostic robustness and flicker-free temporal consistency.

\begin{figure*}[t]
    \centering
    \includegraphics[width=0.95\textwidth]{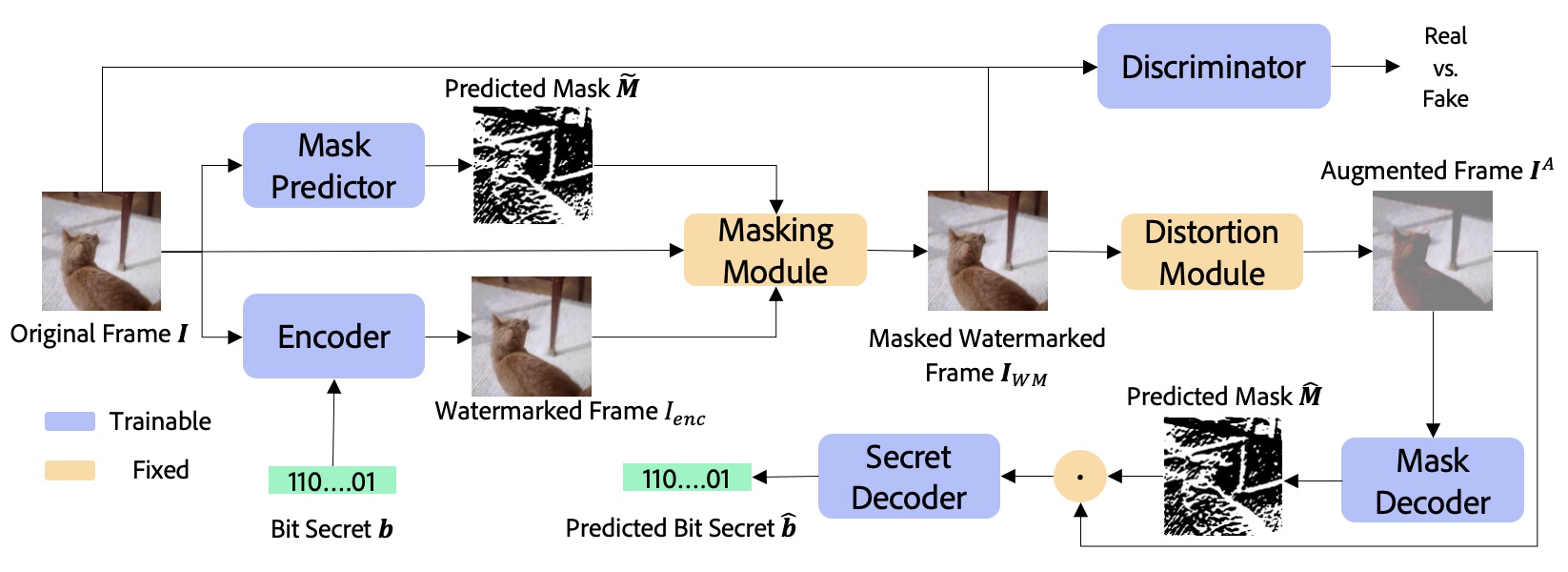}
    \caption{
        \textbf{FlowMark method.} 
        Overview of the FlowMark architecture for mask-guided video watermarking. 
        The \textbf{Encoder} embeds a secret bit sequence \( \vect{b} \) into the input frame \( \vect{I} \), producing a watermarked frame \( \vect{I}_{\text{enc}} \).
        Simultaneously, the \textbf{Mask Predictor} generates a spatial mask \( \tilde{\vect{M}} \) indicating regions suitable for watermark embedding. 
        In the \textbf{Masking Module}, JND-based modulation and dark-region adaptive masking are applied to the embedding residual to improve imperceptibility before spatial masking. Further, it combines \( \vect{I}_{\text{enc}} \) and \( \vect{I} \) using \( \tilde{\vect{M}} \) to produce the masked watermarked frame \( \vect{I}_{\text{wm}} \).  
        A \textbf{Distortion Module} applies differentiable augmentations (e.g., compression, color, or geometric transformations) to obtain the augmented frame \( \vect{I}^A \). The \textbf{Discriminator} provides adversarial supervision to enhance perceptual realism by distinguishing original and watermarked frames. 
        The \textbf{Mask Decoder} reconstructs the mask \( \hat{\vect{M}} \), while the \textbf{Secret Decoder} recovers the embedded message \( \hat{b} \) from the masked augmented frame \( \vect{I}_{\text{mask}} \). 
        Blue blocks denote trainable components; yellow blocks represent fixed differentiable operators.
        }
    \label{fig:method}
\end{figure*}

\noindent\textbf{Media Provenance.}
Media provenance aims to ensure the authenticity and traceability of digital content by recording its origin, authorship, and transformation history.  
The C2PA standard~\cite{c2pa} formalizes this concept through cryptographically signed metadata manifests that travel with media assets to support transparency and attribution.  
Recent research has extended provenance to domains such as archival integrity~\cite{archangel1,archangel2}, AI training consent~\cite{decorait}, and licensing frameworks~\cite{balan2023ekila,contentarcs,accct,jpegtrust2025}.  
However, metadata alone is easily stripped through screenshots, re-encoding, or social media compression.  
To maintain persistent identity, content-aware methods such as perceptual hashing~\cite{nguyen2021,Zhang2020manip,Bharati2021} and watermarking~\cite{devi2009,baba2009,weng2019high,Collomosse2024,Petrov-NeurIPS-2025} are increasingly used to re-link assets with their provenance manifests.  
Our work extends this vision by exploring watermark-based provenance, enabling authenticity tracking within video media.

\section{Method}
\label{sec:method}

\minisection{Overview}
Given an input frame \(\vect{I} \in \mathbb{R}^{H \times W \times 3}\) and a binary watermark message \(\vect{b} \in \{0,1\}^{l}\), 
the model produces a watermarked frame \(\vect{I}_W\) while ensuring reliable watermark recovery under compression and geometric distortions. 
Our framework consists of three components (Fig.~\ref{fig:method}): a spatial mask predictor that determines suitable embedding regions, a message-conditioned encoder that inserts the watermark, and sequential decoder branches that reconstruct the embedding mask and message bits after simulated distortions. The architectural details are provided in the supplementary material. 

Further, we introduce (i) a dark-region adaptive residual modulation mechanism that attenuates watermark strength in perceptually sensitive low-luminance regions, (ii) temporal change consistency constraints that preserve frame-to-frame motion dynamics, and (iii) a three-stage progressive training strategy that transitions from clean images to image-level augmentations and finally to real video codec compression. Together, these components yield a watermarking system that achieves high visual fidelity, temporal stability, and strong resilience to modern compression pipelines.

\minisection{Mask Predictor.}The Mask Predictor determines spatial regions suitable for watermark insertion. 
Given an input frame $\vect{I} \in \mathbb{R}^{H \times W \times 3}$, it predicts a soft embedding map $\vect{M}_{\text{soft}} = f_{\text{mask}}(\vect{I}) \in [0,1]^{H \times W},$
where larger values indicate higher embedding preference.
To enable discrete spatial selection while preserving gradient flow, we apply a straight-through estimator (STE) to obtain a binary mask $\tilde{\vect{M}} = \text{STE}(\vect{M}_{\text{soft}}),$
where the forward pass performs hard binarization, and the backward pass propagates gradients unchanged.

To prevent degenerate solutions such as all-zero or all-one masks, we impose a ratio regularization term that constrains the average mask activation toward a target coverage ratio:
\begin{equation}
\mathcal{L}_{\text{mask}} =
\lambda_{\text{mask}} \cdot \frac{1}{B}
\sum_{b=1}^{B}
\left(
\bar{m}_b - \rho
\right)^2,
\end{equation}
where $\rho$ is the target ratio of activated pixels, $\bar{m}_b$ denotes the mean value of $\tilde{\vect{M}}$ for sample $b$ in a batch of size $B$.

\minisection{Encoder.}
The Encoder \(f_{\text{enc}}(\vect{I}, \vect{b}; \theta_e)\) embeds the binary watermark \(\vect{b}\) into the input frame \(\vect{I}\), producing a watermarked output \(\vect{I}_{\text{wm}}\). 
It learns a latent watermark representation that preserves perceptual quality while enabling reliable bit recovery.

The encoder follows a U-Net-like architecture with skip connections to preserve fine structural details while enabling semantic conditioning through a message projection branch. 
The binary message is first projected into a spatial feature representation and concatenated with the input image before being processed by the backbone network. 
The encoder predicts an intermediate watermark-enhanced frame \(\vect{I}_{\text{enc}}\).

To maintain imperceptibility, we adopt the Just Noticeable Difference (JND) modulation strategy from prior mask-based watermarking approaches~\cite{sander2025wam, hu2025maskimagewatermarking, fernandez2024video}. 
The watermarking residual is adaptively scaled according to local luminance and contrast sensitivity, where $\mu$ controls the global embedding strength. In addition to JND modulation, we apply a luminance-adaptive residual scaling mechanism to suppress visible artifacts in perceptually sensitive dark regions. 
Let $L(x,y)$ denote the per-pixel luminance computed using the BT.601 formulation. 
We define a soft dark-region mask
\begin{equation}
\vect{D}(x,y) =
d_{\min} + (1 - d_{\min}) 
\sigma\big( (L(x,y) - \tau)s \big),
\end{equation}
where $\tau$ is a luminance threshold, $s$ controls transition sharpness, and $d_{\min}$ enforces a minimum watermark strength. 
The final watermarked frame is then multiplied with the predicted mask  by mask predictor and dark-region mask as given by:
\begin{equation}
\vect{I}_{\text{wm}} = 
\vect{I} + \mu \cdot \tilde{\vect{M}} \odot \vect{D} \odot JND(\vect{I}_{\text{enc}} - \vect{I}).
\end{equation}
This luminance-aware modulation reduces watermark strength in low-intensity regions while preserving embedding capacity in perceptually resilient areas.

During training, FlowMark is optimized using a combination of fidelity, perceptual, spatial and temporal objectives. 
A pixel-level reconstruction term enforces similarity between the watermarked and original frames:
\begin{equation}
\mathcal{L}_{\text{pix}} =
\frac{1}{N} \sum_{i=1}^{N}
\left\| \vect{I}_{\text{wm}}^{(i)} - \vect{I}^{(i)} \right\|_2^2.
\end{equation}

To further improve perceptual quality, we use a perceptual similarity term:
\begin{equation}
\mathcal{L}_{\text{lpips}} = LPIPS(\vect{I}_{\text{wm}}, \vect{I}),
\end{equation}
which penalizes visually noticeable differences using deep feature representations.

Additionally, we employ GAN-based adversarial training with a BCE objective and gradient penalty from TrustMark~\cite{bui2025trustmark} to improve the watermarked image quality:
\begin{equation}
\mathcal{L}_{\text{GAN}} =
\text{BCE}(D(x_{\text{real}}), 1)
+
\text{BCE}(D(x_{\text{fake}}), 0)
+
\lambda_{\text{GP}}
\mathbb{E}_{\hat{x}}
\left[
(\|\nabla_{\hat{x}} D(\hat{x})\|_2 - 1)^2
\right],
\end{equation}
where $D(\cdot)$ denotes the discriminator, $x_{\text{real}}$ and $x_{\text{fake}}$ are original and watermarked frames respectively, and $\hat{x} = t x_{\text{real}} + (1-t)x_{\text{fake}}$ with $t \sim \mathcal{U}(0,1)$. The term $\lambda_{\text{GP}}$ controls the strength of the gradient penalty.

In preliminary experiments, we observed that unconstrained watermark embedding often produces grid-like or salt-and-pepper residual patterns, corresponding to spatially scattered high-frequency perturbations, resulting in grid-like artifacts. 
To mitigate this behavior and encourage spatial smoothness, we regularize the watermark residual 
\(\vect{R} = \vect{I}_{\text{wm}} - \vect{I}\)
using spatial total variation loss:
\begin{equation}
\mathcal{L}_{\text{rtv}} =
\frac{1}{HW}
\sum_{x,y}
\left(
\left| R_{x+1,y} - R_{x,y} \right|
+
\left| R_{x,y+1} - R_{x,y} \right|
\right).
\end{equation}

Finally, to explicitly reduce temporal flicker introduced by frame-wise watermark embedding and to preserve motion continuity across video frames, we introduce a Temporal Change MSE loss. 
Let $\{\vect{I}^t\}_{t=1}^{T}$ and $\{\vect{I}^t_{\text{wm}}\}_{t=1}^{T}$ denote the original and watermarked video sequences. 
Defining $\Delta \vect{I}^t = \vect{I}^{t+1} - \vect{I}^t$, and $\Delta \vect{I}_{\text{wm}}^t = \vect{I}_{\text{wm}}^{t+1} - \vect{I}_{\text{wm}}^t$,
we minimize the discrepancy between temporal changes:
\begin{equation}
\mathcal{L}_{\text{tcm}} =
\lambda_{\text{tcm}}
\left\|
\Delta \vect{I}^t_{\text{wm}}
-
\Delta \vect{I}^t
\right\|_2^2.
\end{equation}
By jointly enforcing spatial smoothness and temporal consistency, the encoder learns watermark representations that are both perceptually coherent and motion-consistent across frames.
The overall encoder objective is defined as $\mathcal{L}_{\text{enc}} =
\lambda_{\text{pix}} \mathcal{L}_{\text{pix}}
+ \lambda_{\text{lpips}} \mathcal{L}_{\text{lpips}}
+ \lambda_{\text{rtv}} \mathcal{L}_{\text{rtv}}
+ \lambda_{\text{tcm}}\mathcal{L}_{\text{tcm}}+\lambda_{\text{GAN}}\mathcal{L}_{\text{GAN}}.$

\minisection{Distortion Module}
To simulate real-world degradations encountered during online video sharing, FlowMark employs the differentiable noise and distortion framework introduced in \textit{VideoSeal}~\cite{fernandez2024video}. The watermarked frame $\vect{I}_{\text{wm}}$ is passed through a stochastic distortion operator $\mathcal{D}(\cdot)$ to produce the augmented frame $\vect{I}_A = \mathcal{D}(\vect{I}_{\text{wm}})$. The operator randomly applies a combination of valuemetric transformations (e.g., brightness, contrast, hue, saturation and blur), compression transformations (e.g., JPEG, H.$264$, H.$264$ RGB, and H.$265$) and geometric transformations (e.g., cropping, rotation, perspective, resizing, and flipping), effectively simulating post-processing and platform-specific compression artifacts.  
(See Supplementary for the complete list of transformations and parameter ranges.)

\minisection{Secret and Mask Decoder}
The decoder in FlowMark performs two tightly coupled tasks: (1) reconstructing the spatial mask that identifies the watermark embedding regions, and (2) extracting the binary watermark message from the masked and degraded input. Each task is supervised by a dedicated loss to ensure accurate recovery under compression and distortion.

Given the degraded watermarked frame $\vect{I}^{A} = \mathcal{D}(\vect{I}_{\text{wm}})$, the decoder first predicts a spatial mask $\hat{\vect{M}} = f_{\text{dec-mask}}(\vect{I}^{A}; \theta_d),$
corresponding to the embedding regions. The objective is to reconstruct the original binary mask $\tilde{\vect{M}}$ used during watermark insertion. A pixel-wise mean squared error loss enforces spatial consistency between predicted and ground-truth masks:
\begin{equation}
\mathcal{L}_{\hat{\text{M}}} =
\frac{1}{H W}
\sum_{h=1}^{H} \sum_{w=1}^{W}
\left(
\hat{M}_{h,w} - \tilde{M}_{h,w}
\right)^2.
\end{equation}

Once the spatial mask has been reconstructed, it is applied to the distorted frame to isolate watermark-bearing regions $\vect{I}_{\text{mask}} = \vect{I}^{A} \odot \hat{\vect{M}}$.
The masked frame is then passed through the bit extraction branch to predict the embedded watermark message $\hat{\vect{b}}=f_{\text{dec-sec}}(\vect{I}_{mask}; \theta_d)$.

The predicted bits are supervised using a mean squared error loss:
\begin{equation}
\mathcal{L}_{\text{bit}} =
\frac{1}{l}
\sum_{j=1}^{l}
\left(
\hat{b}_j - b_j
\right)^2.
\end{equation}

The overall decoder objective is defined as $\mathcal{L}_{\text{dec}} =
\lambda_{\text{bit}}\mathcal{L}_{\text{bit}} + \lambda_{\hat{\text{M}}}\mathcal{L}_{\hat{\text{M}}}.$

\begin{figure}[t]
    \centering
    \includegraphics[width=\columnwidth]{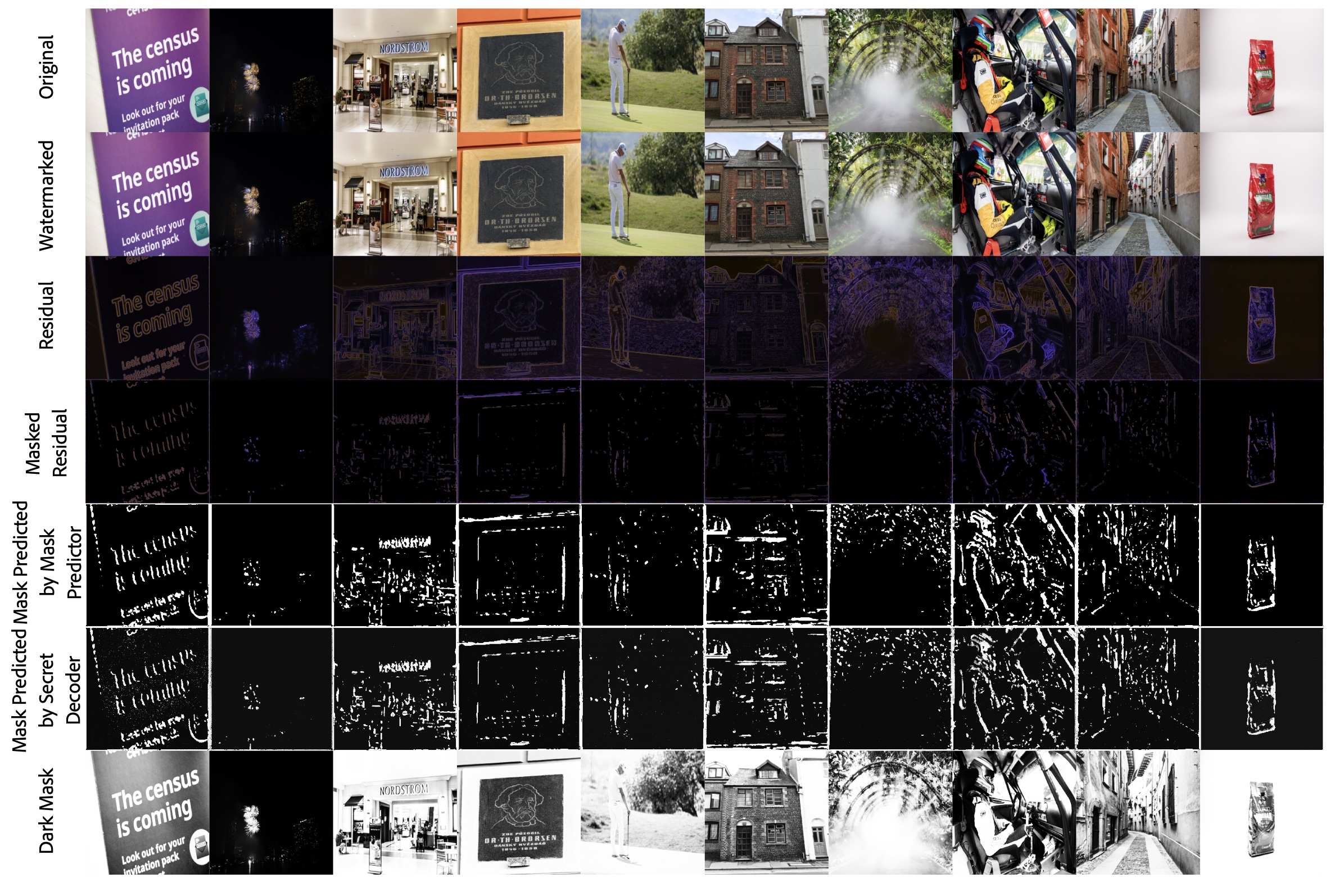}
    \caption{
        Visualization of watermark embedding and mask prediction on SA-1b images. 
From top to bottom: original frames $\vect{I}$, watermarked frames $\vect{I}^{wm}$, full residual \(R = \vect{I}^{enc} - \vect{I}\), masked residual , mask predicted by the Mask Predictor, mask reconstructed by the Secret Decoder, and the luminance-based dark mask. 
The residual maps illustrate that watermark energy is concentrated in perceptually resilient regions rather than uniformly distributed. 
The predicted masks exhibit content-adaptive spatial localization, while the dark mask attenuates embedding strength in low-luminance areas to reduce visible artifacts. 
Together, these visualizations demonstrate spatially coherent, perceptually adaptive, and temporally stable watermark embedding.
    }
    \label{fig:residual_masked_vs_unmasked}
    \vspace{-4mm}
\end{figure}

\subsection{Training Strategy}
FlowMark is trained using a progressive three-stage curriculum designed to improve robustness while preserving perceptual quality and temporal stability. The training pipeline gradually increases distortion severity, allowing stable optimization under increasingly realistic distribution conditions.

\minisection{Stage 1: Clean Pretraining.}
In the first stage, the model is trained without distortion ($\mathcal{D} = \text{Identity}$). 
This stage allows the mask predictor, encoder, and decoder to learn stable embedding and recovery under ideal conditions.

\minisection{Stage 2: Image-Level Robustness.}
In the second stage, image-level augmentations are introduced, including photometric transformations, geometric transformations and resolution changes. 
Dark-region adaptive masking is enabled during this stage to improve perceptual robustness under compression-prone conditions. 
The mask predictor is frozen to stabilize spatial embedding behavior, while the encoder and decoder adapt to moderate distortions.

\minisection{Stage 3: Video Codec Robustness.}
In the final stage, non-differentiable video codec compression (e.g., H.264, H.265, VP9, AV1) is introduced with progressively increasing severity. 
The mask predictor is still frozen and the distortions are applied clip-wise to simulate realistic distribution pipelines. 

This staged curriculum enables FlowMark to transition from clean embedding to real-world robustness in a stable and controlled manner.
Across all three stages, the model is optimized using the joint objective that combines encoder, decoder, and mask prediction losses: $\mathcal{L}_{\text{total}} =
\mathcal{L}_{\text{mask}} + \mathcal{L}_{\text{enc}}
+ \mathcal{L}_{\text{dec}}.$

\subsection{Structured Bit Embedding for Video Authenticity}
Unlike prior methods such as VideoSeal~\cite{fernandez2024video}, which embed a single message repeatedly across \(k\) video frames, FlowMark adopts a structured bit-embedding format enabling frame-level traceability and improved authenticity verification. Each frame carries a unique bit sequence \(\vect{b} \in \{0,1\}^{128}\) partitioned as $\vect{b} = [\vect{b}_{\text{vid}} \, || \, \vect{b}_{\text{frm}}],$
where \(\vect{b}_{\text{vid}} \in \{0,1\}^{112}\) encodes a persistent \textit{video identifier} and \(\vect{b}_{\text{frm}} \in \{0,1\}^{16}\) represents the \textit{frame index}.  

This design allows FlowMark to embed both video-level provenance and frame-level ordering directly in the watermark. The \(112\)-bit video ID remains constant across frames, while the \(16\)-bit frame ID increments sequentially to enable tamper detection such as frame insertion or reordering. In contrast, methods like VideoSeal mitigate flicker by repeating identical messages, which limits frame-level verification. Through this structured bit design and flow-consistent embedding, FlowMark achieves verifiable authenticity with high visual fidelity and temporal smoothness.

\definecolor{headergray}{gray}{0.85}
\definecolor{mygray}{gray}{0.88}

\begin{table*}[t]
\centering
\small
\setlength{\tabcolsep}{4.8pt}
\renewcommand{\arraystretch}{1.25}
\caption{Comparison of watermarking fidelity, robustness, and accuracy under identity condition across SA-1b (image) and SA-V (video) datasets. [KEYS: Acc.: Accuracy]}
\vspace{-3mm}
\begin{adjustbox}{width=\textwidth}
\begin{tabular}{l|c|ccccc|cccccc}
\hline\hline
\rowcolor{headergray}
 & Bit & \multicolumn{5}{c|}{\textbf{SA-1b (Image Dataset)}} & \multicolumn{6}{c}{\textbf{SA-V (Video Dataset)}} \\[-0.4ex]
\arrayrulecolor{headergray}\specialrule{0.7pt}{0pt}{0pt}
\arrayrulecolor{black}
\rowcolor{headergray}
\multirow{-2}{*}{\textbf{Method}} & 
Len. &
\textbf{Bit Acc. (↑)} & \textbf{SSIM (↑)} & \textbf{LPIPS (↓)} & \textbf{PSNR (↑)} & \textbf{Mask Acc. (↑)} &
\textbf{Bit Acc. (↑)} & \textbf{SSIM (↑)} & \textbf{LPIPS (↓)} & \textbf{PSNR (↑)} & \textbf{VMAF (↑)} & \textbf{Mask Acc. (↑)} \\
\hline
HiDDeN~\cite{zhu2018hidden}     & $32$  & $1.00$ & $0.927$ & $0.229$ & $30.36$ & $-$ & $0.99$ & $0.857$ & $0.362$ & $30.19$ & $74.61$ & $-$ \\ 
\rowcolor{mygray}
MBRS~\cite{mbrs}        & $\textbf{128}$ & $0.99$ & $0.997$ & $0.003$ & $45.60$ & $-$ & $1.00$ & $0.995$ & $0.008$ & $46.55$ & $94.10$ & $-$ \\ 
CIN~\cite{ma2022towards}        & $\textbf{128}$ & $1.00$ & $0.997$ & $0.019$ & $44.90$ & $-$ & $1.00$ & $0.994$ & $0.032$ & $45.80$ & $92.93$ & $-$ \\ 
\rowcolor{mygray}
TrustMark~\cite{bui2025trustmark}  & $100$ & $1.00$ & $0.995$ & $\textbf{0.002}$ & $42.09$ & $-$ & $1.00$ & $0.995$ & $\textbf{0.003}$ & $43.07$ & $89.36$ & $-$ \\ 
WAM~\cite{sander2025wam}         & $32$ & $1.00$ & $0.989$ & $0.031$ & $39.86$ & $0.98$ & $1.00$ & $0.981$ & $0.047$ & $40.72$ & $89.78$ & $0.95$ \\  
\rowcolor{mygray}
MaskWM~\cite{hu2025maskimagewatermarking}  & $\textbf{128}$ & $1.00$ & $0.983$ & $0.022$ & $43.19$ & $\textbf{1.00}$ & $1.00$ & $0.952$ & $0.018$ & $45.98$ & $88.26$ & $\textbf{0.99}$ \\ 
VideoSeal~\cite{fernandez2024video} & $96$ & $0.99$ & $0.999$ & $0.009$ & $47.39$ & $-$ & $0.99$ & $\textbf{0.998}$ & $0.013$ & $48.02$ & $93.77$ & $-$ \\
\rowcolor{headergray}
\textbf{FlowMark} & $\textbf{128}$ & $\mathbf{1.00}$ & $\mathbf{0.999}$ & $\mathbf{0.002}$ & $\mathbf{49.08}$ & ${0.95}$ &
$\mathbf{1.00}$ & $\mathbf{0.998}$ & ${0.011}$ & $\mathbf{50.41}$ & $\mathbf{98.95}$ & ${0.96}$ \\ 
\hline\hline
\end{tabular}
\end{adjustbox}
\vspace{0mm}
\label{tab:identity_fidelity_flow}
\vspace{-5mm}
\end{table*}
\section{Experiments}
\vspace{-3mm}
\subsection{Implementation Details}
\vspace{-2mm}
\minisection{Metrics}
We evaluate both imperceptibility and robustness across image and video datasets. Performance is measured using Bit Accuracy (\(\uparrow\)), the fraction of correctly recovered watermark bits, and Mask Accuracy (\(\uparrow\)), which assesses how precisely the model predicts spatial embedding regions.  
Imperceptibility is quantified using perceptual metrics: SSIM (\(\uparrow\)) for structural similarity, PSNR (\(\uparrow\)) for pixel-level fidelity, and LPIPS (\(\downarrow\)) for learned perceptual distance. For videos, we additionally report VMAF (\(\uparrow\))~\cite{netflix2019video}, a perceptual quality metric that correlates closely with human judgment.

\minisection{Training Details}
We train FlowMark on the Adobe Stock video dataset and evaluate on SA-1B~\cite{kirillov2023segment} and SA-V following the protocol of VideoSeal~\cite{fernandez2024video}. From SA-1B, 500k images are randomly sampled and resized to $256\times256$. For video evaluation, we use the SA-V test set comprising $155$ videos at $24$ fps. During training, all frames are resized to $256\times256$, and clips of length 16 are sampled every $10$ frames. Training follows a three-stage curriculum, stage $1$ running for 100k steps, and stages $2$ and $3$ for $200k$ steps. Stage $1$ is trained with learning rate $1\times10^{-4}$, Stage $2$ with $5\times10^{-5}$, and Stage $3$ with $2\times10^{-5}$, all using AdamW with a cosine scheduler. Training is conducted on $8$ NVIDIA A100 80GB GPUs with batch size $6$ per GPU. Noise distortions are disabled in Stage $1$, introduced after $5k$ steps in Stage $2$, and after $3k$ steps in Stage $3$ with a progressive ramp schedule. Unless otherwise specified, we use message length $l=128$, target mask ratio $\rho=0.5$, and loss weights $\lambda_{\text{mask}}= 25$, $\lambda_{\text{pix}}= 1.0 $, $\lambda_{\text{lpips}}= 0.3$, $\lambda_{\text{rtv}}= 2.0$, $\lambda_{\text{tcm}}= 0.5$, $\lambda_{\text{bit}}= 20$,$\lambda_{\text{GAN}}= 0.5$, and $\lambda_{\hat{\text{M}}}= 0.5$. We set $\tau = 0.3$, $d_{\min} = 0.2$, $s = 10$, $\mu = 1.0$, and $\rho = 0.5$ in all experiments.

\begin{figure}[t]
    \centering
    \includegraphics[width=1\columnwidth]{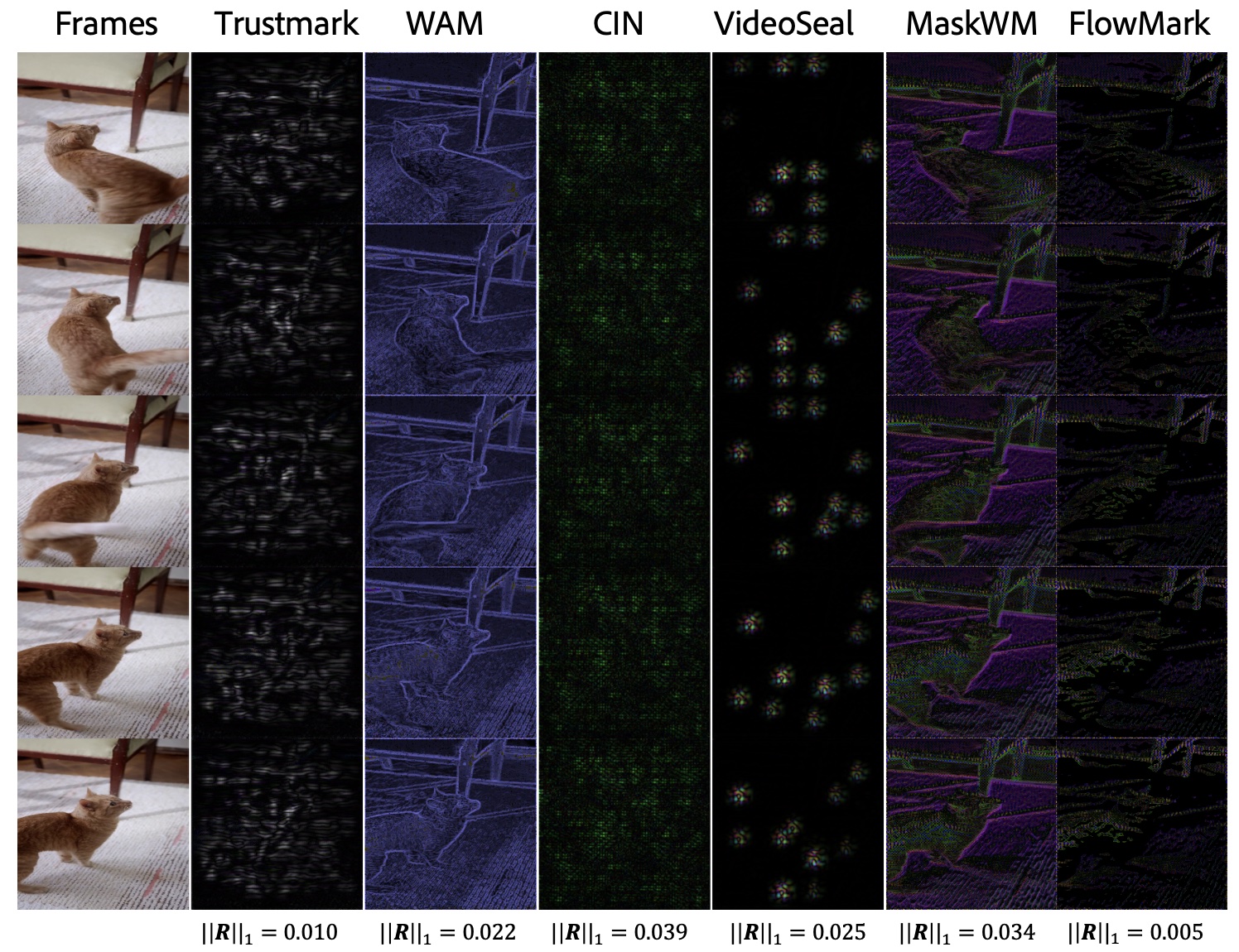}
    \caption{
        Visualization of frame-wise residuals ($\vect{R} = \vect{I} - \vect{I}_{\text{wm}}$) across methods. FlowMark produces the lowest residual magnitude ($||\vect{R}||_{1}$), indicating minimal perceptual distortion and the most stable watermark embedding across frames.
    }
    \label{fig:residual}
    \vspace{-6mm}
\end{figure}

\vspace{-3mm}
\subsection{Comparison With Baselines}
As FlowMark operates at the frame level, we compare it against recent image and video watermarking frameworks, including HiDDeN \cite{zhu2018hidden}, MBRS \cite{mbrs}, CIN \cite{ma2022towards}, TrustMark \cite{bui2025trustmark}, WAM \cite{sander2025wam}, and VideoSeal \cite{fernandez2024video}. While these models achieve strong image-level robustness, they fail to maintain temporal consistency when extended to video, often leading to flicker or loss of frame-level traceability. 

Across both SA-1b (image) and SA-V (video) benchmarks (Table~\ref{tab:identity_fidelity_flow}), FlowMark achieves unmatched performance in watermark fidelity, robustness, and perceptual quality. It consistently attains perfect bit accuracy while maintaining the highest PSNR of \(49.08\,\text{dB}\) (images) and \(50.41\,\text{dB}\) (videos), with near-perfect SSIM scores (\(>0.99\)), a combination no prior approach attains. Even high-capacity methods such as MBRS and CIN show visible degradation at similar payloads, whereas FlowMark preserves pristine image quality. The highest VMAF score of \(98.95\) further confirms that FlowMark produces videos visually indistinguishable from their originals in standard streaming conditions.  

\cref{fig:residual_masked_vs_unmasked} and~\cref{fig:residual} illustrate how FlowMark achieves this advantage.~\cref{fig:residual_masked_vs_unmasked} shows the predicted masks and masked residuals, demonstrating that FlowMark embeds signals adaptively within high-texture or motion-stable regions. The residuals are further diminished using the predicted mask.~\cref{fig:residual} visualizes frame-wise residuals (\(\vect{R} = \vect{I} - \vect{I}_{\text{wm}}\)) across methods, FlowMark exhibits the lowest residual magnitude (\(\|\vect{R}\|_1 = 0.005\)), reflecting minimal perceptual distortion and superior temporal stability.

\definecolor{headergray}{gray}{0.85}
\definecolor{mygray}{gray}{0.88}

\begin{table}[t]
\centering
\small
\setlength{\tabcolsep}{5pt}
\renewcommand{\arraystretch}{1.25}
\caption{Comparison of bit accuracy under four robustness conditions (Valuemetric, Geometric, Compression, and Combined) across SA-1b (image) and SA-V (video) datasets. FlowMark achieves consistently high bit accuracy across all perturbations, indicating strong robustness.}
\vspace{-2mm}
\begin{adjustbox}{width=\columnwidth}
\begin{tabular}{l|cccc|cccc}
\hline\hline
\rowcolor{headergray}
 & \multicolumn{4}{c|}{\textbf{SA-1b (Image Dataset)}} & \multicolumn{4}{c}{\textbf{SA-V (Video Dataset)}} \\[-0.4ex]
\arrayrulecolor{headergray}\specialrule{0.7pt}{0pt}{0pt}
\arrayrulecolor{black}
\rowcolor{headergray}
\multirow{-2}{*}{\textbf{Method}} &
\textbf{Valuemetric} & \textbf{Geometric} & \textbf{Compression} & \textbf{Combined} &
\textbf{Valuemetric} & \textbf{Geometric} & \textbf{Compression} & \textbf{Combined} \\
\hline
HiDDeN~\cite{zhu2018hidden}     & $0.88$ & $0.76$ & $1.00$ & $0.70$ & $0.88$ & $0.68$ & $0.83$ & $0.61$ \\ 
\rowcolor{mygray}
MBRS~\cite{mbrs}   & $0.95$ & $0.52$ & $0.99$ & $0.50$ & $0.89$ & $0.50$ & $0.79$ & $0.50$ \\ 
CIN~\cite{ma2022towards}       & $0.91$ & $0.52$ & $1.00$ & $0.50$ & $0.93$ & $0.50$ & $0.90$ & $0.49$ \\ 
\rowcolor{mygray}
TrustMark~\cite{bui2025trustmark}  & $0.98$ & $0.65$ & $1.00$ & $0.53$ & $0.93$ & $0.60$ & $0.87$ & $0.51$ \\ 
WAM~\cite{sander2025wam}        & $1.00$ & $0.81$ & $1.00$ & $0.86$ & $0.92$ & $0.85$ & $0.86$ & $0.73$ \\  
\rowcolor{mygray}
MaskWM~\cite{hu2025maskimagewatermarking}  & $1.00$ & $0.85$ & $1.00$ & $0.83$ & $0.89$ & $0.71$ & $0.80$ & $0.69$ \\
VideoSeal~\cite{fernandez2024video}   & $0.93$ & $0.83$ & $0.99$ & $0.91$ & $0.90$ & $0.85$ & $0.85$ & $0.73$ \\ 
\rowcolor{headergray}
\textbf{FlowMark} & $\mathbf{0.99}$ & $\mathbf{0.89}$ & $\mathbf{1.00}$ & ${\textbf{0.92}}$ 
& $\mathbf{0.98}$ & $\mathbf{0.88}$ & ${\bf0.88}$ & $\mathbf{0.80}$ \\ 
\hline\hline
\end{tabular}
\end{adjustbox}
\label{tab:robustness_comparison_no_identity}
\vspace{-3mm}
\end{table}

FlowMark achieves the highest VMAF score ($98.95$) on the SA-V dataset, indicating superior perceptual video quality under compression. Unlike competing methods that sacrifice visual fidelity for robustness, FlowMark maintains high spatial quality (PSNR/SSIM) while preserving video-level perceptual realism (VMAF), demonstrating coherent, motion-consistent watermark embedding across frames.

\vspace{-3mm}
\subsection{Robustness Results}

\begin{figure}[t]
    \centering
    \includegraphics[width=0.95\columnwidth]{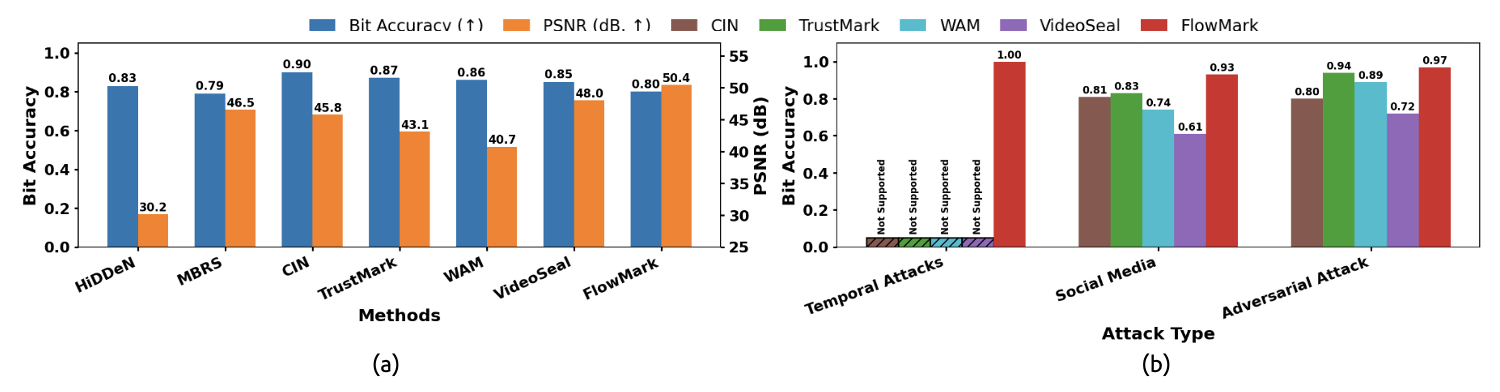}
    \caption{
        \textbf{Robustness performance  on the SA-V video dataset.}
        (a) Comparison of bit accuracy and PSNR across watermarking methods under video compression evaluation using FFMpeg. (b) Robustness under temporal edits, social media compression, and adversarial attacks. 
    }
    \label{fig:video_comparison}
    \vspace{-3mm}
\end{figure}

\begin{figure*}[t]
    \centering
    \includegraphics[width=0.7\linewidth]{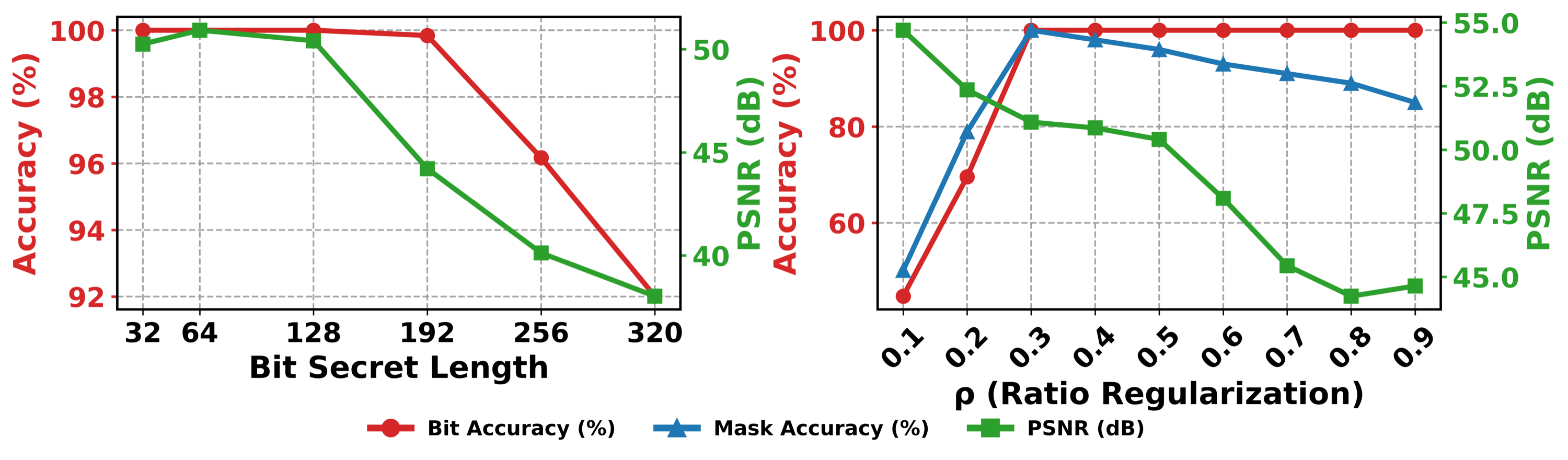}
    \vspace{-4mm}
    \caption{Ablation studies on (left) bit secret length and (right) mask ratio regularization.}
    \label{fig:ablation_bitlength}
    \vspace{-6mm}
\end{figure*}

Table~\ref{tab:robustness_comparison_no_identity} reports bit accuracy under four robustness settings: valuemetric, geometric, compression, and combined distortions (details in supplement), on SA-1b (image) and SA-V (video) datasets. FlowMark consistently achieves the highest accuracy across all perturbations, showing strong resilience to brightness, scaling, rotation, and compression artifacts. On SA-$1$b, it attains \(0.99\) under valuemetric, \(0.89\) under geometric, and a perfect \(1.00\) under compression distortions, yielding an overall combined accuracy of \(0.92\). While MaskWM performs competitively on valuemetric and compression tests, FlowMark remains markedly more stable under geometric transformations where other methods degrade sharply.

On SA-V, FlowMark maintains near-perfect robustness with accuracies of \(0.98\), \(0.88\), and \(0.88\) for valuemetric, geometric, and compression distortions, respectively, outperforming all baselines, especially in geometric robustness where MBRS and CIN drop below \(0.60\). These results confirm that FlowMark’s content-adaptive masking and flow-consistent embedding preserve both fidelity and stability under real-world video degradations.

Figure~\ref{fig:video_comparison} presents quantitative comparisons under realistic video degradation scenarios. In (a), we evaluate bit accuracy and PSNR under video compression by locally re-encoding videos using FFMPEG with multiple CRF settings and codecs (details in supplementary). Each video is decoded and re-saved at varying CRF levels to simulate practical compression pipelines. FlowMark maintains strong bit accuracy while preserving high PSNR across CRF levels. 

In (b), we evaluate robustness under temporal edits, social media compression, and adaptive adversarial attacks. 
For temporal evaluation, FlowMark uses its structured per-frame embedding strategy, where each frame carries a unique frame index within a 128-bit message. 
This enables explicit detection of frame swap ($10-50\%$), frame drop ($10-50\%$), frame insertion ($10-50\%$), temporal resampling, and frame interpolation through ordering violations or frame-count inconsistencies. 
FlowMark achieves $100\%$ tamper detection accuracy while preserving perfect decoding on non-manipulated frames. 
Manipulated frames are correctly flagged rather than falsely decoded. 
The average frame-count error (500 frames per video) is $0$ (swap), $0$ (drop), $4$ (insertion), $8$ (interpolation), and $7$ (resampling), demonstrating precise temporal traceability under video edits. 
In contrast, prior methods use their respective embedding schemes (e.g., repeated or random watermarks across frame windows), which do not encode frame-level ordering information and therefore cannot detect temporal manipulations.

To assess social media robustness, we upload and download $20$ watermarked videos from the SA-V validation set via YouTube and Facebook, and evaluate decoding performance after platform compression. FlowMark maintains high bit accuracy under these real-world distribution pipelines.
For adaptive adversarial attacks, we evaluate VAE-based reconstruction attacks (Stable Diffusion VAE; Ballé~\etal ICLR 2018; Cheng~\etal CVPR 2020). 
FlowMark is fine-tuned for $5$k steps on 8 NVIDIA A$100$ GPUs under these distortions and achieves $97.17\%$ decoding accuracy, demonstrating resilience under adaptive threat models.

\vspace{-3mm}
\subsection{Ablations}

\minisection{Effect of Bit Secret Length.}
We evaluate the effect of bit length on robustness and perceptual quality. As shown in Fig.~\ref{fig:ablation_bitlength}~(left), FlowMark maintains nearly perfect accuracy and over \(50\,\text{dB}\) PSNR up to \(128\) bits, indicating imperceptible embedding. Beyond this, performance drops as spatial capacity saturates, at \(192\) and \(320\) bits, accuracy falls to about \(94\%\) and PSNR to \(42\,\text{dB}\). This trade-off highlights the balance between information density and perceptual fidelity in localized watermarking. We adopt \(128\)-bit embedding as the default, balancing robustness, fidelity, and recovery.

\minisection{Effect of Mask Ratio Regularization (\(\rho\)).}
The ratio parameter \(\rho\) controls the proportion of pixels used for embedding the watermark. As shown in Fig.~\ref{fig:ablation_bitlength}~(center), small ratios (\(\rho < 0.2\)) restrict watermark coverage, reducing bit and mask recovery.  Larger ratios extend masking into smooth regions, which are less tolerant to noise, lowering perceptual quality. 

\definecolor{headergray}{gray}{0.85}
\definecolor{mygray}{gray}{0.88}

\begin{table}[t]
\centering
\small
\setlength{\tabcolsep}{5pt}
\renewcommand{\arraystretch}{1.2}
\caption{Ablation study of removing individual FlowMark components and losses.}
\vspace{-4mm}
\begin{adjustbox}{width=1.0\columnwidth}
\begin{tabular}{l|ccccccccccc}
\hline\hline
\rowcolor{headergray}
\textbf{Metric} 
& Baseline 
& Dark Mask 
& JND 
& Mask Pred. 
& Distortion 
& Mask Dec. 
& $L_{\text{mask}}$ 
& $L_{\hat{\text{M}}}$ 
& $L_{\text{rtv}}$ 
& $L_{\text{tcm}}$ 
& $L_{\text{lpips}}$ \\
\hline
Robust Bit Acc. (\%) $\uparrow$
& \(0.80\) & \(0.86\) & \(0.82\) & \(0.81\) & \(0.53\) & \(0.62\) & \(0.71\) & \(0.74\) & \(0.78\) & \(0.79\) & \(0.82\) \\
PSNR (dB) $\uparrow$
& \(50.41\) & \(40.18\) & \(38.35\) & \(38.11\) & \(47.65\) & \(41.18\) & \(42.65\) & \(38.17\) & \(31.11\) & \(37.19\) & \(35.98\) \\
\hline\hline
\end{tabular}
\end{adjustbox}
\vspace{-8mm}
\label{tab:flowmark_ablation}
\end{table}

\minisection{Architecture and Losses Ablation}
Table~\ref{tab:flowmark_ablation} evaluates the contribution of individual components and loss terms in FlowMark relative to the full baseline model (\(0.80\) robust accuracy, \(50.41\)~dB PSNR).
Disabling the proposed dark-region adaptive masking improves robustness from \(0.80\) to \(0.86\), while reducing PSNR to \(40.18\)~dB due to stronger residual attenuation in perceptually sensitive regions. This confirms that luminance-aware modulation enhances fidelity under distortion at the cost of slight robustness reduction.
Disabling JND scaling reduces PSNR significantly to \(38.35\)~dB, highlighting the importance of perceptual weighting for maintaining visual quality. Similarly, removing the Mask Predictor degrades PSNR to \(38.11\)~dB, demonstrating that adaptive spatial localization is essential for visually stable watermark placement.
Excluding the Distortion Module causes robust accuracy to collapse from \(0.80\) to \(0.53\) despite relatively high PSNR (\(47.65\)~dB), indicating severe overfitting to clean data and poor generalization under compression or geometric noise. Removing the Mask Decoder also substantially reduces robustness to \(0.62\), confirming the importance of reconstructing embedding regions during decoding.

We further analyze the impact of individual loss terms used in our framework. Removing the mask loss (\(L_{\text{mask}}\)) lowers robustness to \(0.71\), while excluding the mask reconstruction loss applied on the secret decoder(\(L_{\hat{\text{M}}}\)) reduces robustness to \(0.74\), demonstrating that structured spatial supervision on mask modules improves decoding stability. Disabling residual total variation (\(L_{\text{rtv}}\)) severely degrades PSNR to \(31.11\)~dB, confirming its role in suppressing high-frequency artifacts. Finally, removing temporal consistency (\(L_{\text{tcm}}\)) reduces robustness to \(0.79\) and PSNR to \(37.19\)~dB, highlighting the importance of temporal regularization in video watermarking. Omitting perceptual supervision (\(L_{\text{lpips}}\)) also decreases PSNR to \(35.98\)~dB, underscoring its contribution to visual fidelity.
Overall, the full model achieves the best performance, validating the necessity of both spatial and temporal design components.

\vspace{-3mm}
\section{Conclusion}
\label{sec:conclusion}
We presented FlowMark, a mask-guided video watermarking framework that learns to adapt watermark embedding to optimal spatial regions of video content. By predicting content- and motion-aware masks, FlowMark achieves robust, imperceptible, and temporally coherent watermarking without manual intervention. Our experiments demonstrate that this learned masking strategy effectively minimizes residual distortion, reduces flicker, and sustains high fidelity under compression, geometric distortions, social media re-encoding, and temporal editing operations. FlowMark establishes a new paradigm for video watermarking and enables reliable content provenance and temporal authenticity verification for emerging generative and streaming media pipelines.

\section{Additional Experiments}

\begin{figure*}[t]
\centering
\includegraphics[width=\linewidth]{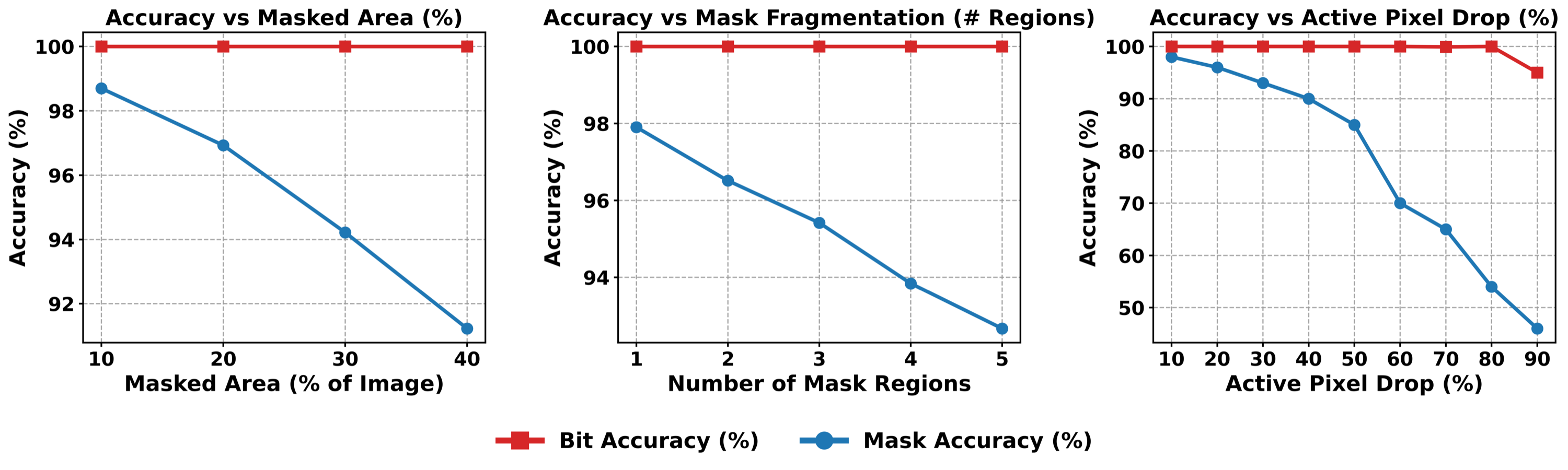}
\vspace{-2mm}
\caption{
\textbf{Ablation on mask sensitivity.}
We evaluate FlowMark under three complementary deletion settings.
(Left) Masked Area Ablation: a contiguous region covering 10--40 percent of the image is removed, and results are averaged over all fragmentation levels (1--5 regions). 
(Middle) Mask Fragmentation Ablation: the total masked area is fixed while the number of disjoint masked regions varies from 1--5, with metrics averaged over the 10--40 percent area levels.
(Right) Active Pixel Drop Ablation: a random percentage of watermark-carrying pixels is deleted.
Across all settings, bit accuracy remains near 100 percent until corruption becomes extreme, while mask accuracy degrades exactly with the loss of spatial coherence in the residual.
}
\label{fig:mask_region_ablation}
\vspace{-3mm}
\end{figure*}

\minisection{Ablation on Mask Sensitivity.}
We evaluate FlowMark under three structured ways of damaging the spatial structure of the watermark to understand how much spatial corruption of the mask can absorb while still preserving message integrity.

First, in the masked‐area setting, we explicitly zero out a contiguous rectangular regions covering between $10-40\%$ of the image and report results averaged over all fragmentation levels ($1-5$ regions). This removes entire blocks of the embedded residual. Bit accuracy stays near $100$ percent because the encoder distributes the message across the entire frame, but mask accuracy decreases steadily since larger deletions remove more of the spatial evidence that the secret decoder relies on.

Second, in the mask‐fragmentation setting, we keep the total removed area fixed but split it into an increasing number of smaller disjoint patches, each averaging over the area levels used in the first setting. Fragmenting the occlusion injects many small breaks into the residual’s spatial continuity. Bit accuracy remains stable, but mask accuracy drops faster than in the previous case because region-wise prediction depends heavily on coherent local structure; many small holes are more disruptive than one large missing region of the same size.

Finally, in the active–pixel drop setting, we simulate highly destructive stochastic noise by randomly deleting a percentage of only the watermark‐carrying pixels. Unlike structured masking, this noise directly attacks the residual itself instead of removing visible content. Bit accuracy stays high until the deletion rate becomes extreme, but mask accuracy collapses rapidly because the fine-grained spatial pattern that the classifier depends on becomes noisy and irregular.

Overall, these ablations show a consistent pattern: FlowMark’s global bitstream is extremely robust to all forms of spatial corruption, whereas region-level accuracy declines exactly in proportion to how much the residual’s spatial structure is disrupted.

\begin{table}[h!]
\centering
\small
\caption{Computational efficiency of embedding and extraction under the unified video-inference setup. Reported GFLOPs reflect the effective compute required following VideoSeal~\cite{fernandez2024video} evaluation, and latency values show the end-to-end CPU/GPU time per second of video~\cite{fernandez2024video}. FlowMark maintains competitive compute cost while achieving practical runtime performance.}
\begin{adjustbox}{width=\columnwidth}
\begin{tabular}{lcccccc}
\hline
\textbf{Method} &
\textbf{Embed GFLOPs} &
\textbf{CPU Encoder (s)} &
\textbf{GPU Encoder (s)} &
\textbf{Extract GFLOPs} &
\textbf{CPU Decoder (s)} &
\textbf{GPU Decoder (s)} \\
\hline
HiDDeN      & $22.4$ & $1.51$ & $0.07$ & $39.0$ & $8.88$ & $0.26$ \\
MBRS        & $32.2$ & $6.70$ & $0.22$ & $27.0$ & $19.98$ & $0.40$ \\
TrustMark   & $10.3$ & $2.24$ & $0.07$ & $4.1$  & $10.39$ & $0.42$ \\
WAM         & $42.6$ & $4.78$ & $0.31$ & $68.7$ & $30.20$ & $1.32$ \\
VideoSeal   & $42.0$ & $9.04$ & $0.35$ & $3.1$  & $26.09$ & $1.11$ \\
FlowMark    & $12.7$ & $2.69$ & $0.09$ & $17.6$ & $15.53$ & $0.70$ \\
\hline
\end{tabular}
\end{adjustbox}
\label{tab:efficiency}
\end{table}

\minisection{Efficiency Analysis.}
Table~\ref{tab:efficiency} compares the computational efficiency of the evaluated watermarking models under a unified video-inference pipeline where each model embeds messages into $256\times256$ frames using temporal propagation with $k=4$, following~\cite{fernandez2024video}. We run the experiment on NVIDIA-A$100$ GPU. The baselines span a wide compute spectrum: HiDDeN~\cite{zhu2018hidden} and MBRS~\cite{mbrs} require relatively high embedding and extraction GFLOPs, TrustMark~\cite{bui2025trustmark} is noticeably lighter, and WAM~\cite{sander2025wam} remains the most expensive due to its attenuation heatmap computation. VideoSeal~\cite{fernandez2024video} shows high embedding cost but unusually low extraction GFLOPs, although its latency remains elevated due to overhead from its compiled execution path. 

FlowMark sits in a balanced region: its embedding GFLOPs fall near the low end while CPU and GPU latency remain competitive. Extraction cost is moderate and its runtime aligns with other practical baselines. In effect, FlowMark avoids the inefficiencies of heavier methods like WAM and MBRS while maintaining stable video-level performance.

\begin{figure*}[t]
    \centering
    \includegraphics[width=1.0\textwidth]{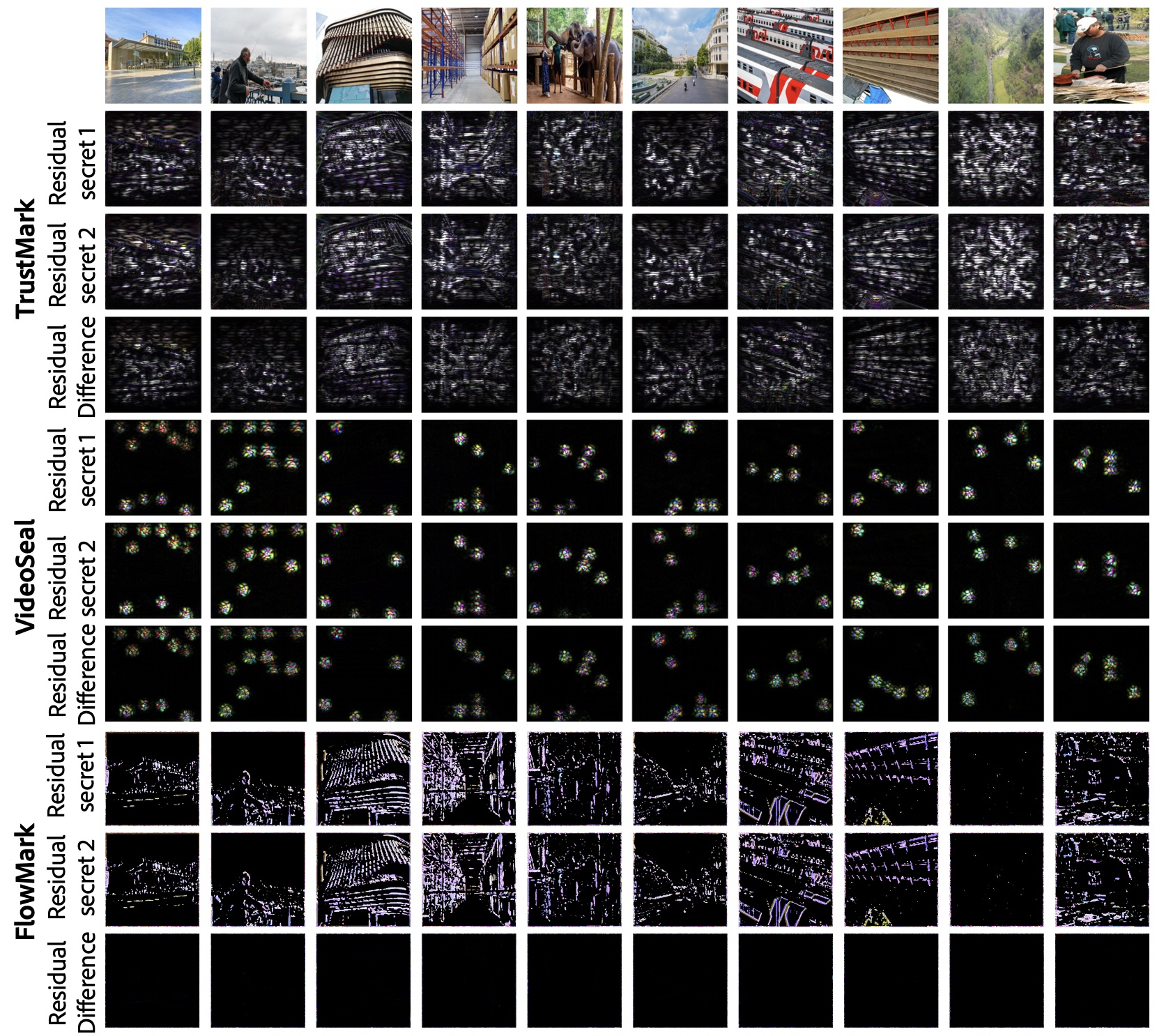}
    \caption{
        Spatial residual comparison across different secrets on SA-I images. Residuals are visually amplified to reveal differences. For each method, we show residuals for two different secrets along with a difference row that computes the pixel-wise difference between the two residuals. TrustMark and VideoSeal exhibit strong variations across secrets, and their difference maps contain large structured artifacts, indicating inconsistent spatial embedding. In contrast, FlowMark produces highly similar residuals across secrets, and its difference maps remain faint and spatially uniform. This consistency reflects FlowMark’s stable spatial encoding and helps minimize frame-to-frame flicker when each video frame is watermarked with a different secret.
        }
    \label{fig:secret_variation}
\end{figure*}

\minisection{Residual Variation With Secret}
~\cref{fig:secret_variation} visualizes how the spatial residual changes when different secrets are embedded into the same SA-I images. For each method, we show two residual maps (one per secret), followed by a difference map computed as the pixelwise absolute difference between the two scaled residuals. All residuals are magnified to make structural variation visible. This setup isolates one question: How much does the embedding pattern change when only the secret changes?

TrustMark shows strong message-dependent variation. Its two residuals differ substantially across secrets, with streaks and high-frequency artifacts shifting position and intensity. The corresponding difference maps display bright, widespread structures, confirming that even small changes in the secret produce large spatial perturbations. In video watermarking, this instability causes frame-to-frame flicker whenever consecutive frames carry different secrets.

VideoSeal exhibits lower but still significant variation. Its residuals contain scattered blob-like activations whose magnitude fluctuate across secrets. The difference maps reveal discrete clusters of change, indicating that although VideoSeal is more stable than TrustMark, its embedding pattern still jitters enough to introduce temporal inconsistency in videos.

FlowMark remains stable across secrets. Its residuals are visually similar, with smooth textures and minimal structural drift. The difference maps show only faint, low-energy patterns, demonstrating that FlowMark effectively decouples message encoding from local pixel structure. This secret-invariant spatial behavior ensures that even when different secrets are embedded in consecutive frames, the residual field stays consistent, producing flicker-free, temporally smooth video watermarking.

\minisection{Full Resolution Watermarking}
FlowMark enables robust watermark embedding at full image resolution while maintaining perceptual quality. Although the model is trained and evaluated at $256 \times 256$ resolution, it generalizes to arbitrary image sizes by upscaling the learned residual and applying luminance-adaptive modulation.  

Given an input image $\mathbf{I} \in \mathbb{R}^{H \times W \times 3}$ and a residual map $\mathbf{R}$ predicted by the encoder, we first upscale the residual to the original resolution, denoted as $\mathbf{R}_{\uparrow}$. 
A luminance-based dark-region mask $\mathbf{D}$ is computed using the BT.601 formulation to attenuate embedding strength in low-intensity regions. 
The final watermarked image is then given by:
\[
\mathbf{I}_W = \mathbf{I} + \mathbf{D} \odot \mathbf{R}_{\uparrow},
\]
where $\odot$ denotes element-wise multiplication. 

This luminance-adaptive residual scaling preserves spatial consistency across resolutions while reducing visible artifacts in perceptually sensitive regions, ensuring robust yet imperceptible full-resolution watermarking.

\begin{figure*}[t]
    \centering
    \includegraphics[width=0.95\textwidth]{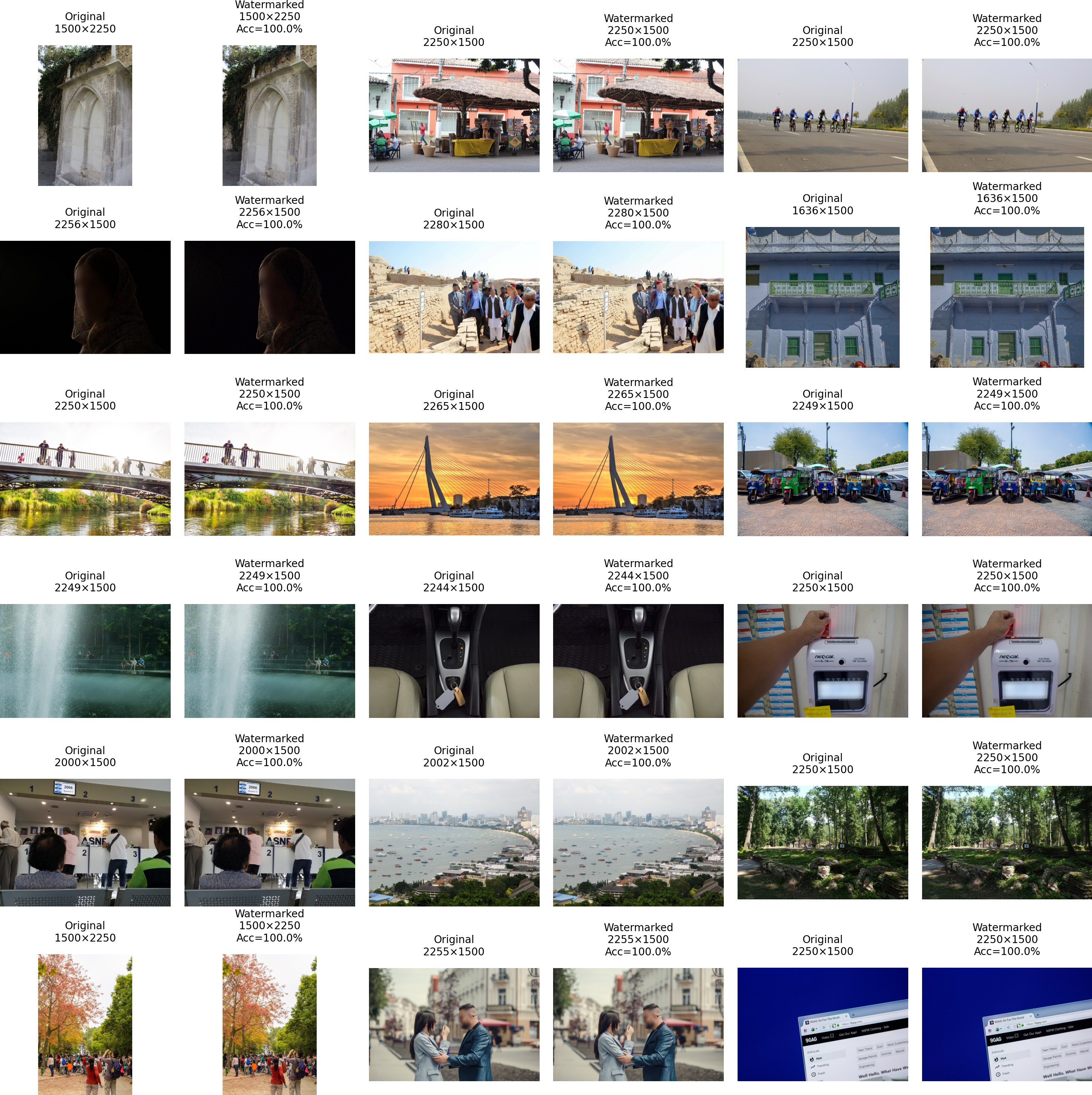}
    \caption{
        \textbf{Full resolution watermarking.}  Each pair shows the original and watermarked images at native resolutions. Residuals are spatially upscaled and blended with the original image to embed the watermark. All samples maintain $100\%$ bit recovery accuracy, with imperceptible visual difference between original and watermarked outputs. 
        }
    \label{fig:full_res}
\end{figure*}

For decoding, the watermarked image $\mathbf{I}_W$ is first downsampled to $256 \times 256$ and passed through the pretrained decoder:
\[
\hat{\mathbf{b}} = \mathcal{D}(\text{resize}(\mathbf{I}_W)),
\]
where $\hat{\mathbf{b}}$ is the recovered bit sequence. This design allows FlowMark to preserve visual fidelity across diverse high-resolution images while maintaining perfect bit recovery, as demonstrated in \cref{fig:full_res}.

\begin{figure*}[t]
    \centering
    \includegraphics[width=\textwidth]{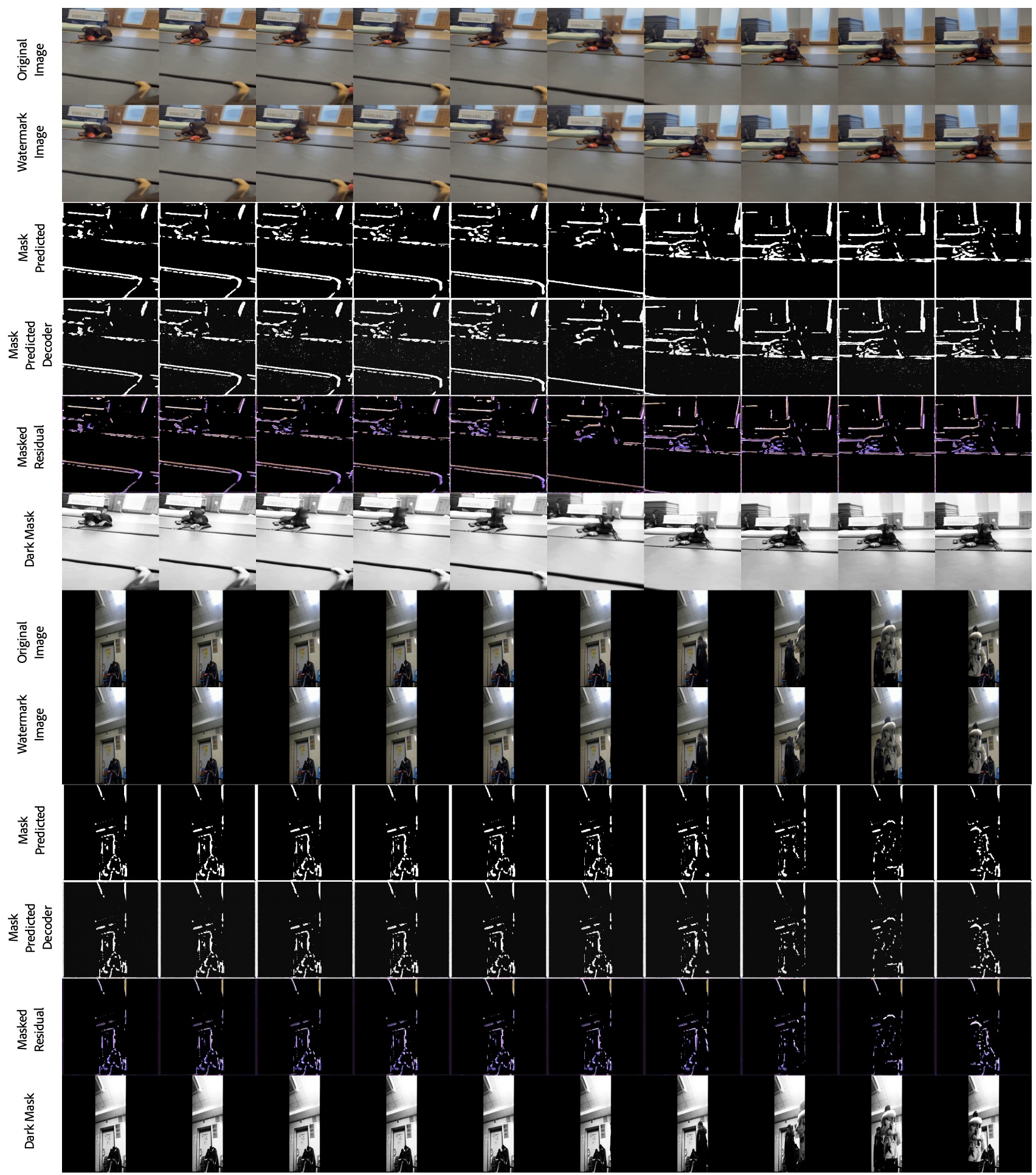}
    \caption{
        \textbf{Visualization of video watermarking across multiple sequences.} Each group shows the original, the watermarked output, the predicted mask from the mask-decoder, the mask predicted by the decoder, the scaled residual between the watermarked and original images, and the dark mask at multiple timestamps. The residuals are scaled for visibility. FlowMark maintains a spatially consistent embedding pattern across frames, resulting in temporally smooth watermarking without flicker.
        }
    \label{fig:video1}
\end{figure*}

\begin{figure*}[t]
    \centering
    \includegraphics[width=1\textwidth]{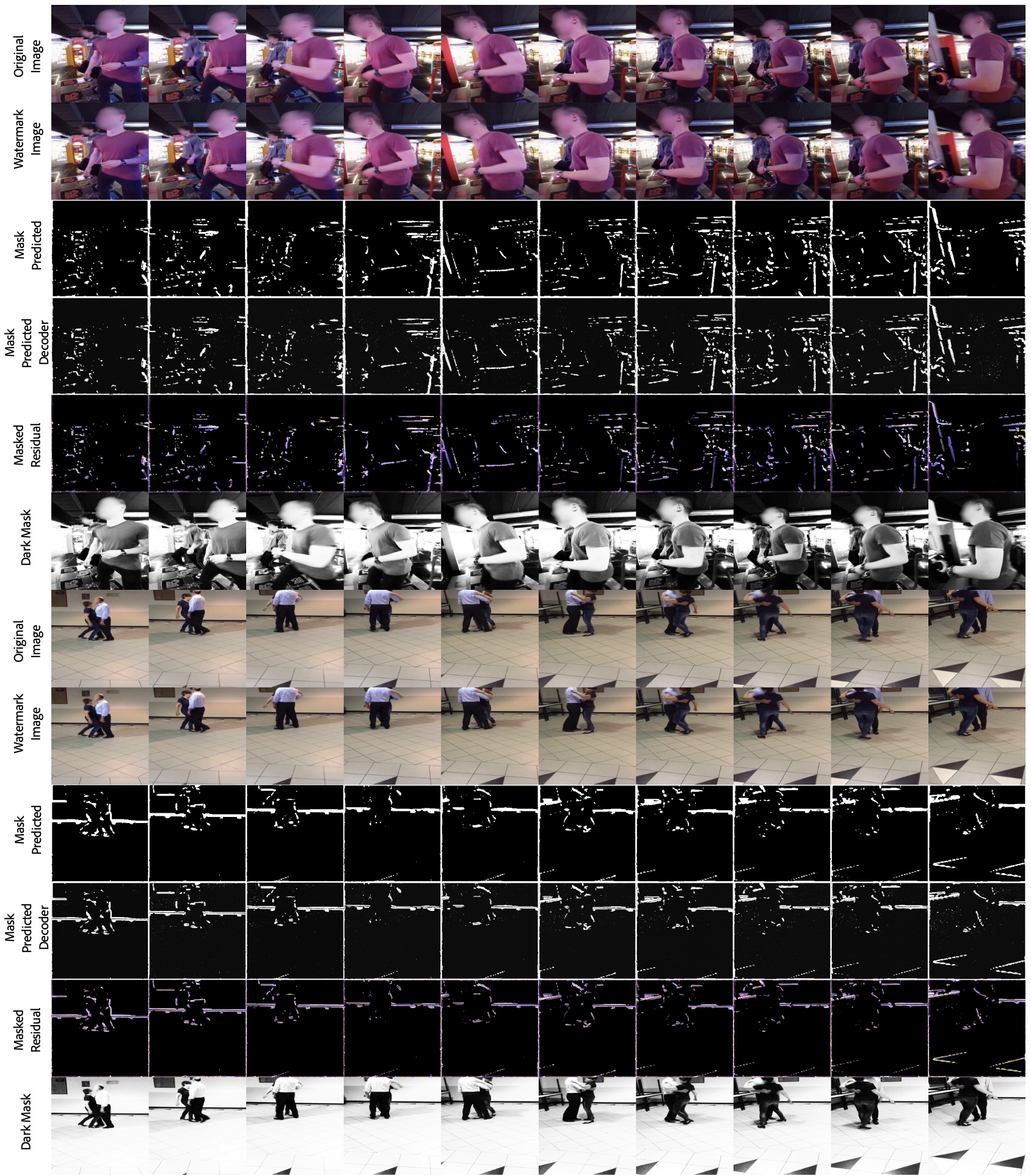}
    \caption{
        \textbf{Visualization of video watermarking across multiple sequences.} Each group shows the original, the watermarked output, the predicted mask from the mask-decoder, the mask predicted by the decoder, the scaled residual between the watermarked and original images, and the dark mask at multiple timestamps. The residuals are scaled for visibility. FlowMark maintains a spatially consistent embedding pattern across frames, resulting in temporally smooth watermarking without flicker.
        }
    \label{fig:video2}
\end{figure*}

\begin{figure*}[t]
    \centering
    \includegraphics[width=1\textwidth]{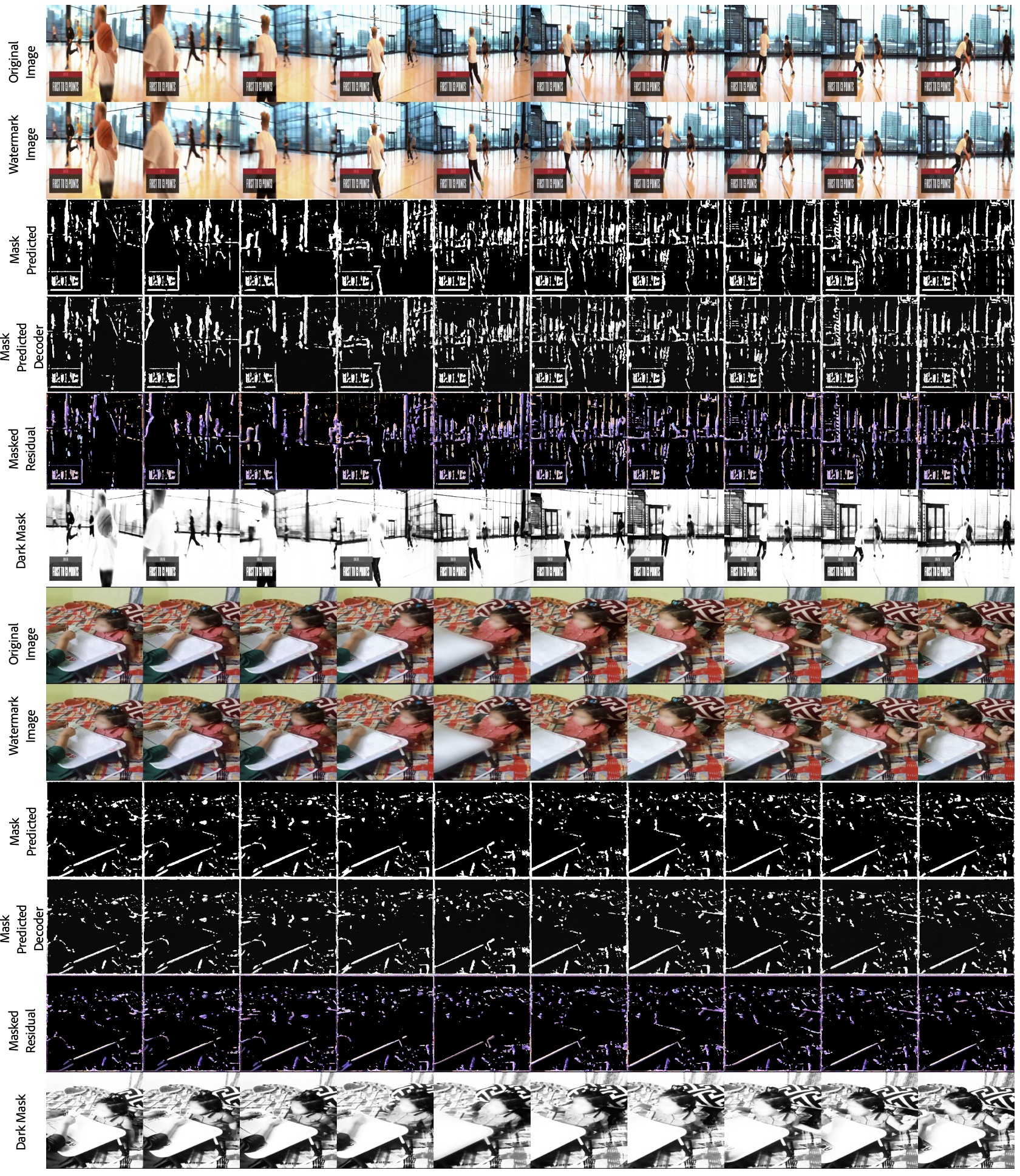}
    \caption{
        \textbf{Visualization of video watermarking across multiple sequences.} Each group shows the original, the watermarked output, the predicted mask from the mask-decoder, the mask predicted by the decoder, the scaled residual between the watermarked and original images, and the dark mask at multiple timestamps. The residuals are scaled for visibility. FlowMark maintains a spatially consistent embedding pattern across frames, resulting in temporally smooth watermarking without flicker.
        }
    \label{fig:video3}
\end{figure*}

\begin{figure*}[t]
    \centering
    \includegraphics[width=1\textwidth]{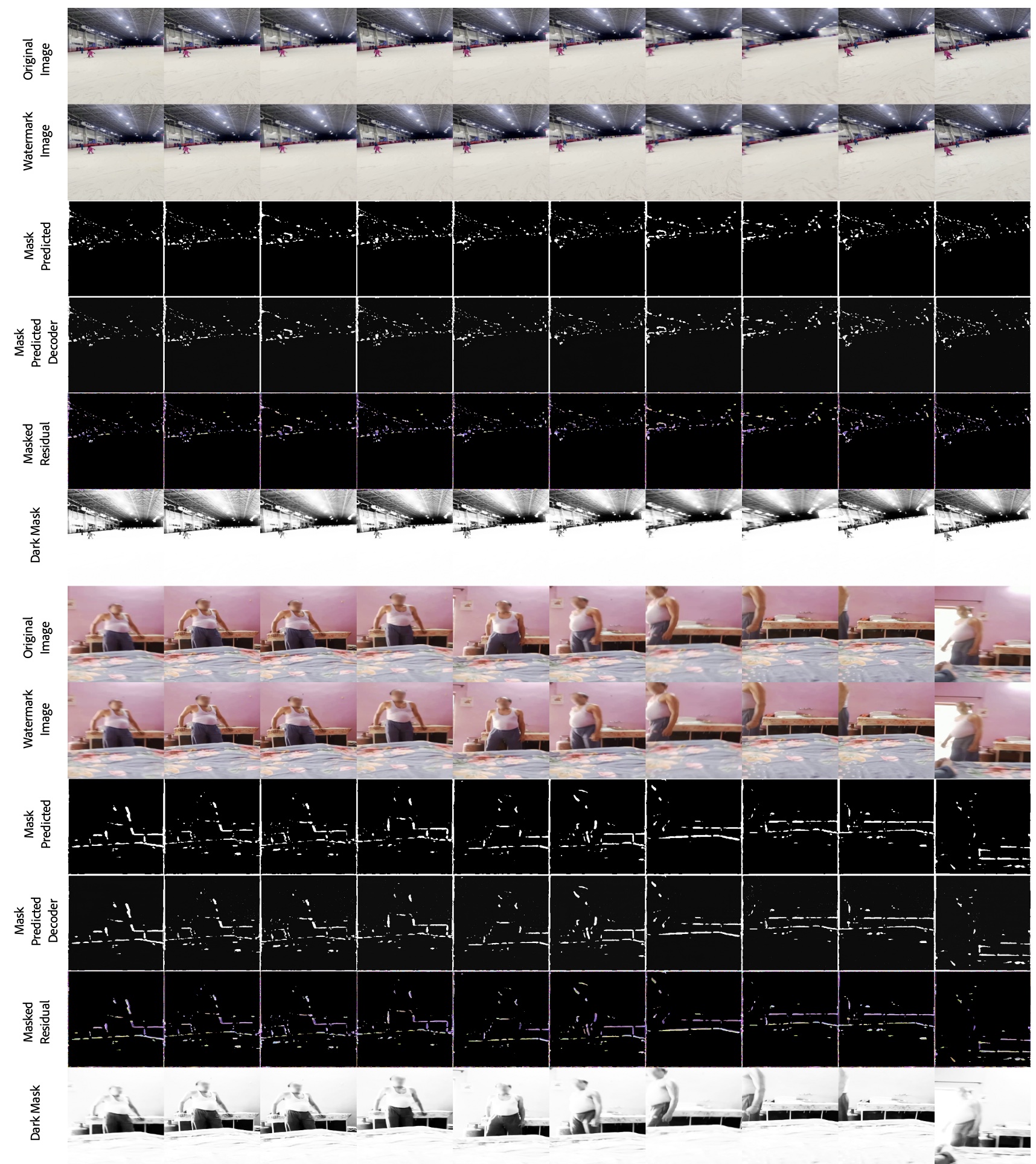}
    \caption{
        \textbf{Visualization of video watermarking across multiple sequences.} Each group shows the original, the watermarked output, the predicted mask from the mask-decoder, the mask predicted by the decoder, the scaled residual between the watermarked and original images, and the dark mask at multiple timestamps. The residuals are scaled for visibility. FlowMark maintains a spatially consistent embedding pattern across frames, resulting in temporally smooth watermarking without flicker.
        }
    \label{fig:video4}
\end{figure*}

\begin{figure*}[t]
    \centering
    \includegraphics[width=1\textwidth]{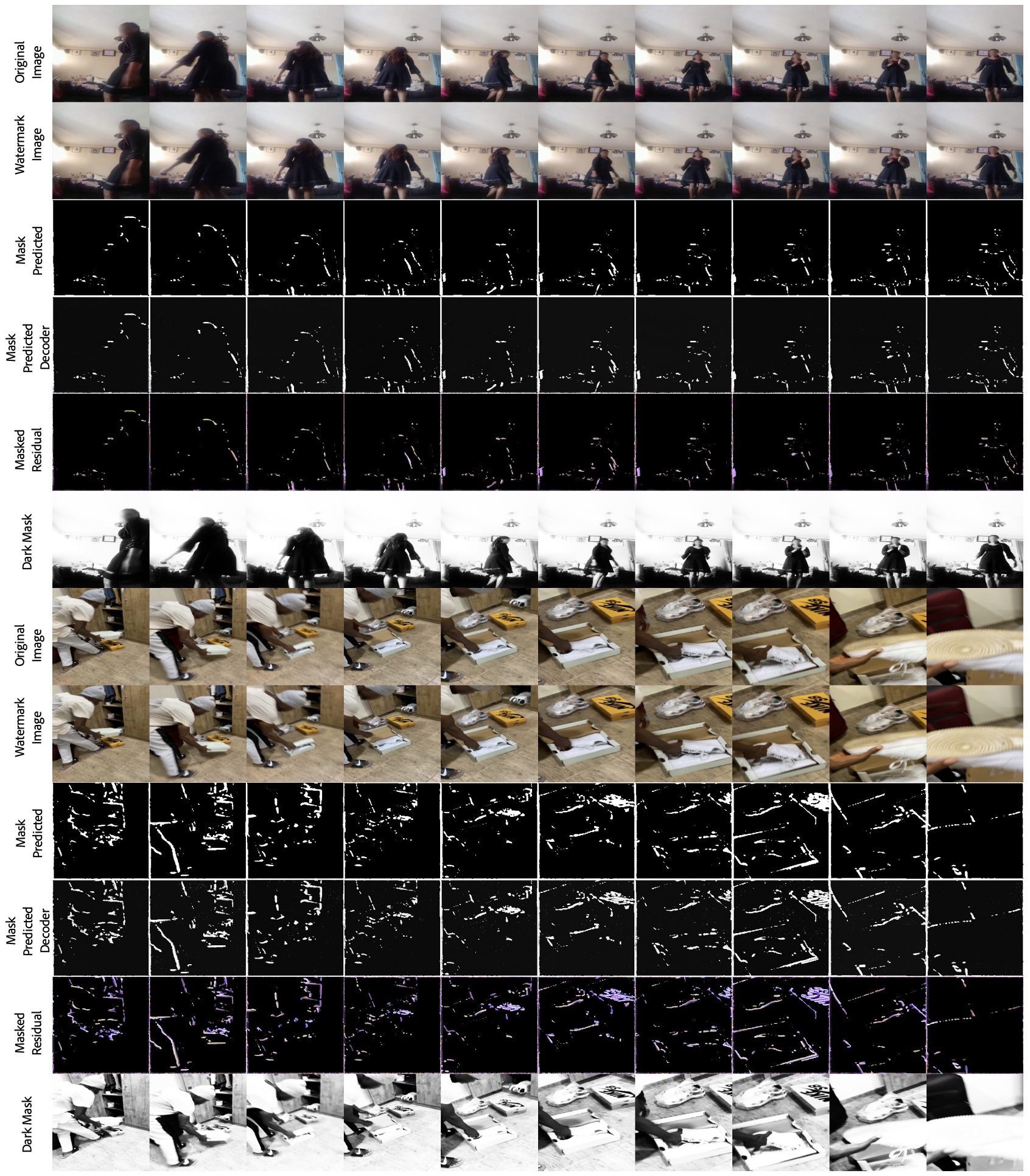}
    \caption{
        \textbf{Visualization of video watermarking across multiple sequences.} Each group shows the original, the watermarked output, the predicted mask from the mask-decoder, the mask predicted by the decoder, the scaled residual between the watermarked and original images, and the dark mask at multiple timestamps. The residuals are scaled for visibility. FlowMark maintains a spatially consistent embedding pattern across frames, resulting in temporally smooth watermarking without flicker.
        }
    \label{fig:video5}
\end{figure*}

\begin{figure*}[t]
    \centering
    \includegraphics[width=1\textwidth]{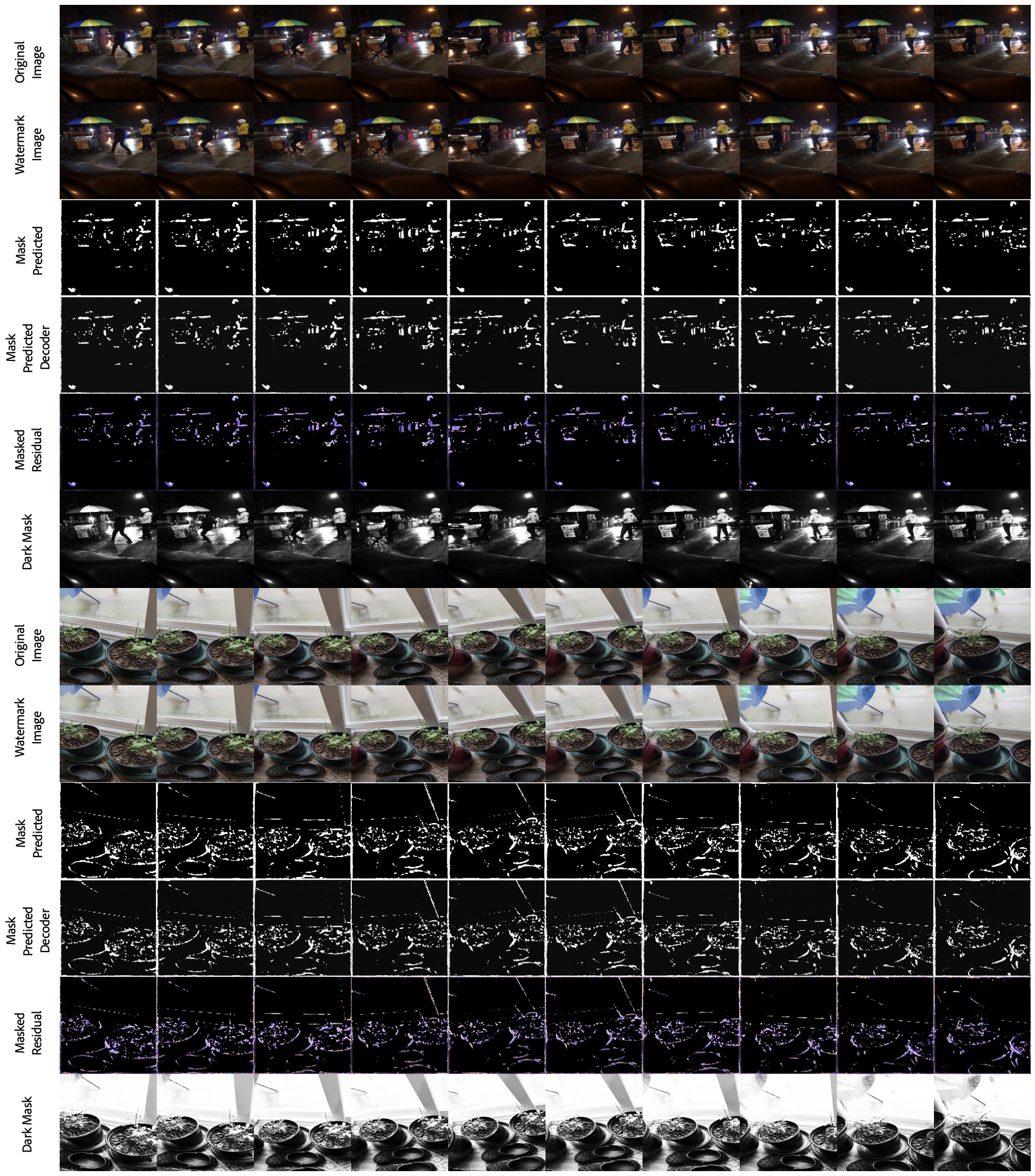}
    \caption{
        \textbf{Visualization of video watermarking across multiple sequences.} Each group shows the original, the watermarked output, the predicted mask from the mask-decoder, the mask predicted by the decoder, the scaled residual between the watermarked and original images, and the dark mask at multiple timestamps. The residuals are scaled for visibility. FlowMark maintains a spatially consistent embedding pattern across frames, resulting in temporally smooth watermarking without flicker.
        }
    \label{fig:video6}
\end{figure*}

\begin{figure*}[t]
    \centering
    \includegraphics[width=1\textwidth]{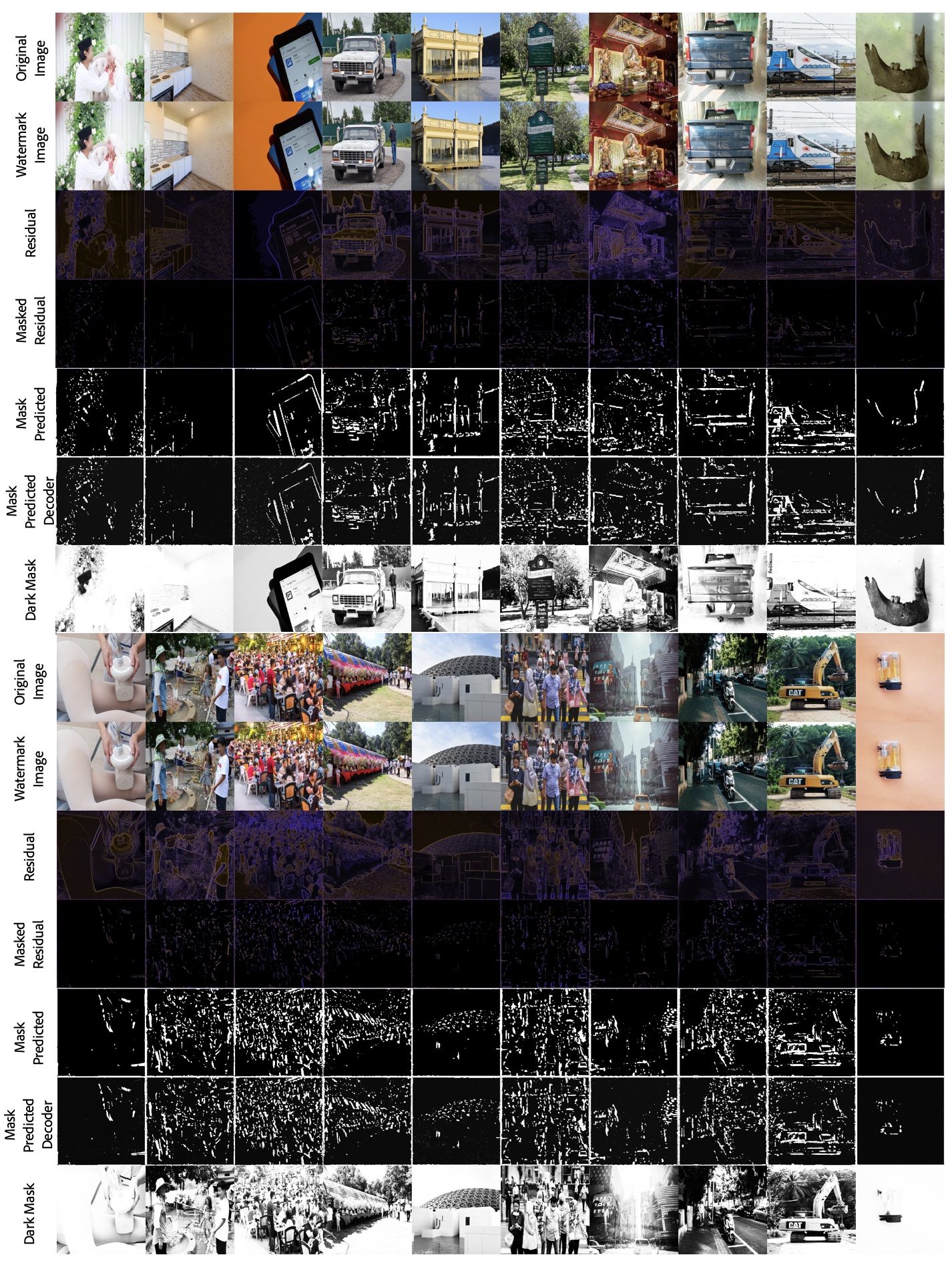}
    \caption{
        \textbf{Visualization of FlowMark on still images.} For each image, we show the original, the watermarked output, the predicted mask from the mask-decoder, the mask predicted by the decoder, the scaled residual between the watermarked and original images, and the dark mask. Residuals are intentionally amplified for visibility. Across diverse image types, FlowMark produces stable, spatially coherent residuals and accurate masks, demonstrating that the watermarking signal remains consistent and minimally intrusive even at full resolution.
        }
    \label{fig:image1}
\end{figure*}

\begin{figure*}[t]
    \centering
    \includegraphics[width=1\textwidth]{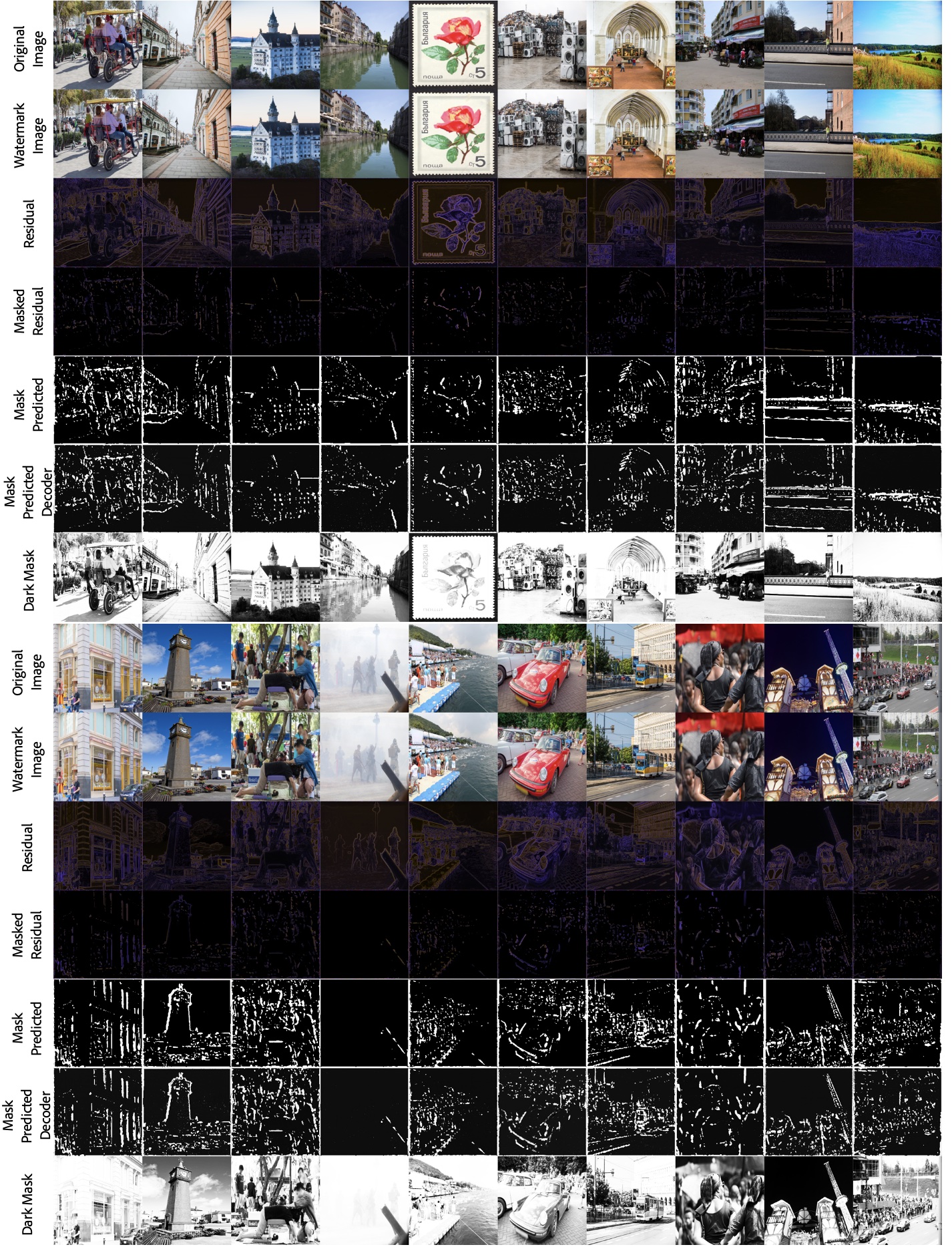}
    \caption{
        \textbf{Visualization of FlowMark on still images.} For each image, we show the original, the watermarked output, the predicted mask from the mask-decoder, the mask predicted by the decoder, the scaled residual between the watermarked and original images, and the dark mask. Residuals are intentionally amplified for visibility. Across diverse image types, FlowMark produces stable, spatially coherent residuals and accurate masks, demonstrating that the watermarking signal remains consistent and minimally intrusive even at full resolution.
        }
    \label{fig:image2}
\end{figure*}

\begin{figure*}[t]
    \centering
    \includegraphics[width=1\textwidth]{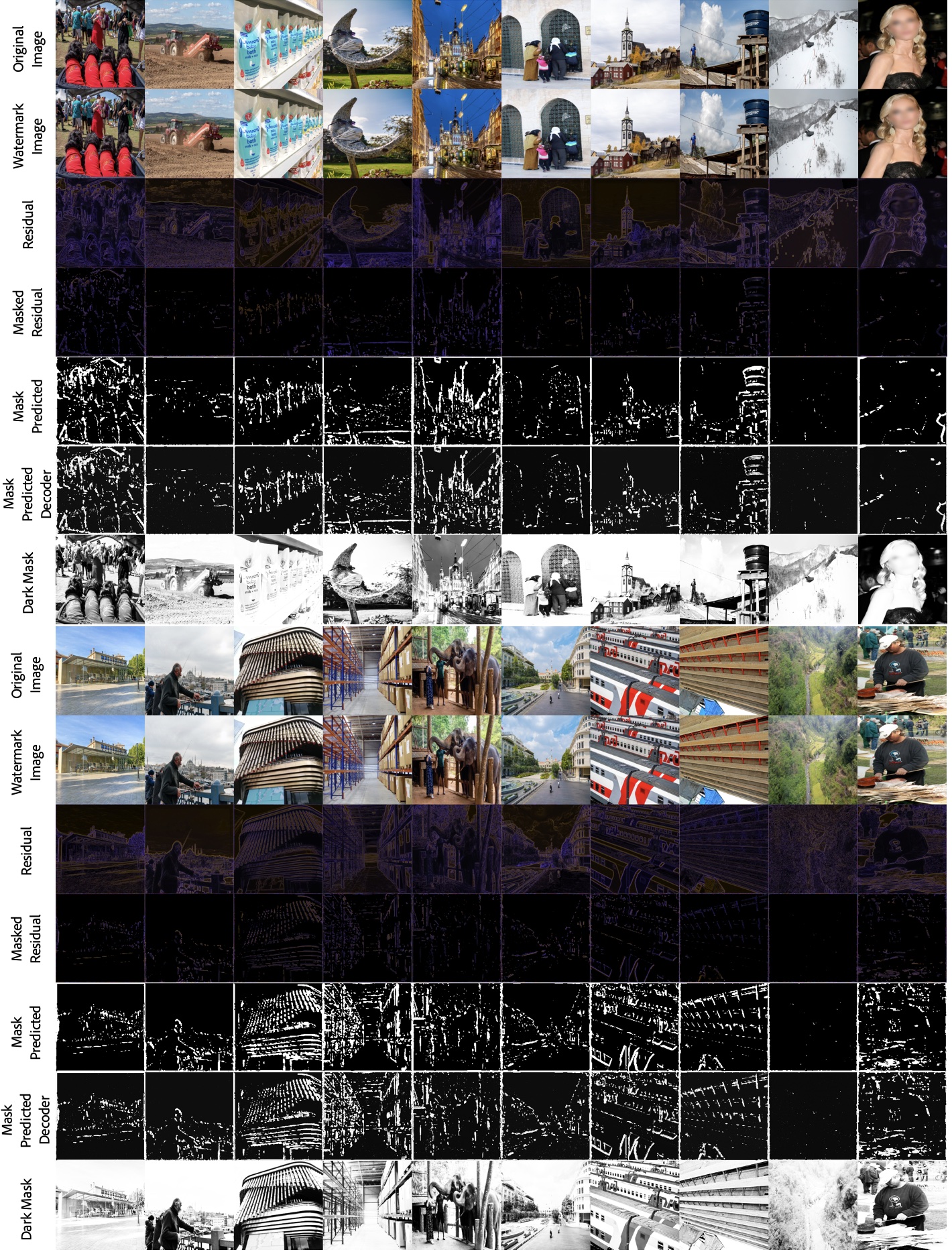}
    \caption{
        \textbf{Visualization of FlowMark on still images.} For each image, we show the original, the watermarked output, the predicted mask from the mask-decoder, the mask predicted by the decoder, the scaled residual between the watermarked and original images, and the dark mask. Residuals are intentionally amplified for visibility. Across diverse image types, FlowMark produces stable, spatially coherent residuals and accurate masks, demonstrating that the watermarking signal remains consistent and minimally intrusive even at full resolution.
        }
    \label{fig:image3}
\end{figure*}

\minisection{Visualization for Images and Video}
FlowMark’s behavior on videos (~\cref{fig:video1},~\cref{fig:video2},~\cref{fig:video3},~\cref{fig:video4}~\cref{fig:video5},~\cref{fig:video6}) and images (~\cref{fig:image1},~\cref{fig:image2},~\cref{fig:image3}) is visualized through the predicted masks, decoder masks, dark masks, and scaled residuals. In videos, the model produces spatially coherent residuals that remain consistent with video frame motion across consecutive frames, even when each frame carries a different secret. Competing approaches such as TrustMark and VideoSeal exhibit large frame-to-frame changes in their residual patterns, creating blocky, drifting, or texture-dependent perturbations. These variations manifest as temporal instability, which appears as flicker when watermarked videos are played back~\cref{fig:secret_variation}. FlowMark avoids this by predicting a stable spatial mask and embedding a residual that remains structurally similar over time, which is exactly what enables smooth video watermarking. We also provide representative qualitative samples for multiple videos in the supplementary zip file having multiple folders, where each folder has all the videos to illustrate these behaviors across diverse scenes, motion patterns, and aspect ratios. The selected examples include both portrait and landscape videos at varying resolutions, highlighting that FlowMark maintains spatial coherence and temporal stability regardless of content type or frame dynamics.

A similar pattern is visible in the image setting. For each still image, FlowMark generates a mask that localizes embedding, and the decoder independently recovers a closely matching mask, indicating strong consistency between the encoder and decoder stages. The residuals (scaled for visibility) remain smooth and spatially coherent across diverse image types, without introducing structured artifacts or input-dependent distortions. Even with amplification, the perturbations look stable and uniform, demonstrating that FlowMark embeds information in a controlled way rather than relying on brittle or image-specific artifacts.

Together, the video and image visualizations show that FlowMark’s residuals and masks are stable, predictable, and highly consistent across consecutive frames and across distinct images. This stability is fundamental: it preserves perceptual quality at full resolution, eliminates temporal flicker in videos, and reliably supports watermark decoding even when running at 256×256 resolution during inference.

\begin{figure}[t]
    \centering
    \includegraphics[width=0.6\linewidth]{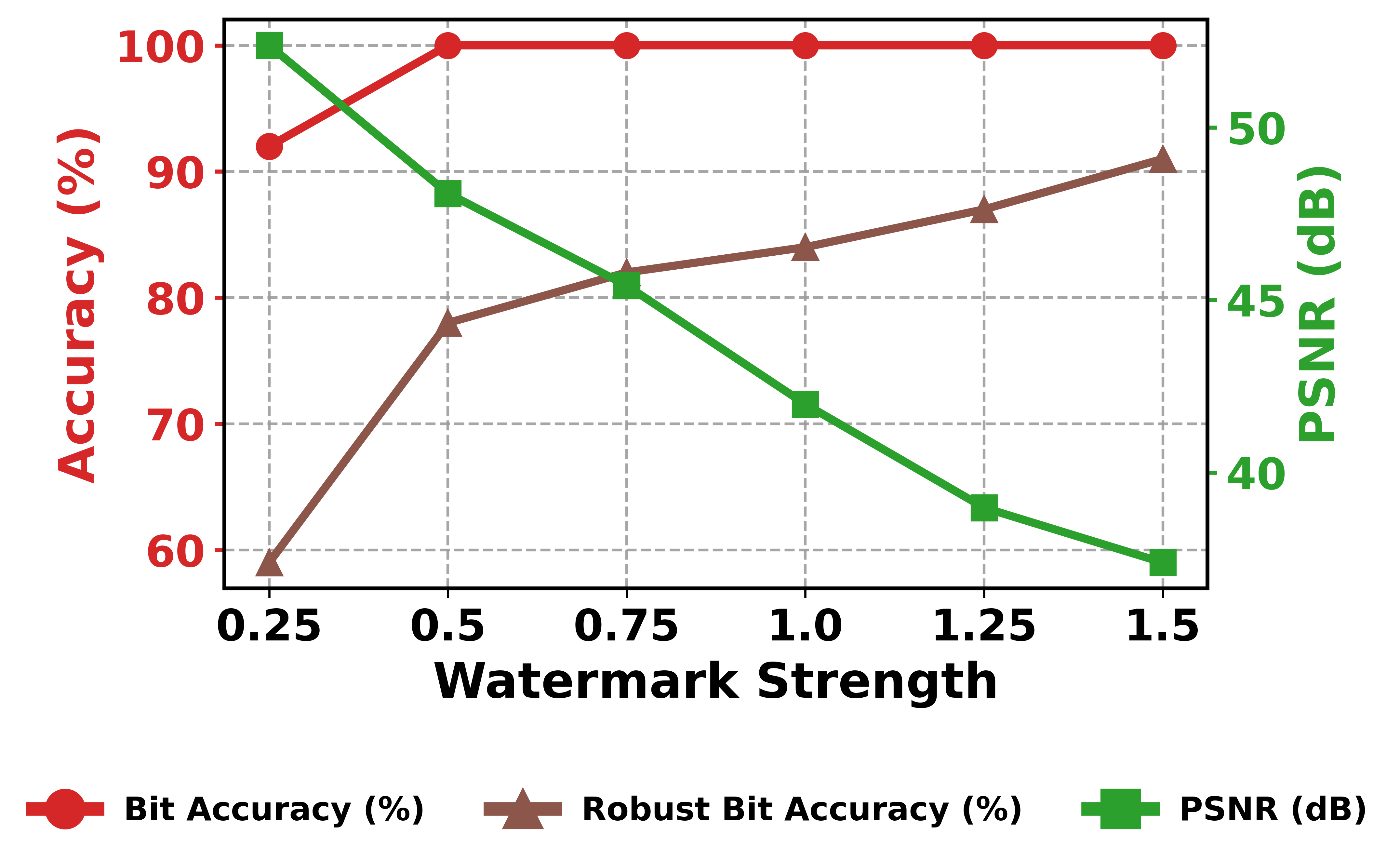}
    \caption{
    \textbf{Watermark strength ablation.}
    Increasing strength improves robustness while reducing PSNR, highlighting the trade-off between robustness and visual quality.
    }
    \label{fig:strength_ablation}
\end{figure}

\minisection{Effect of Watermark Strength.}
Although we utilize a luminance-based dark mask for adaptive strength of the embedded residual, we further study the effect of having a fixed watermark strength, which modulates the amplitude of the residual applied in masked regions during inference. As shown in Fig.~\ref{fig:strength_ablation}, increasing the strength of the embedded signal, improves robustness (combined setting) under different noise augmentations (robustness accuracy rises from \(60\%\) at \(\alpha = 0.25\) to \(92\%\) at \(\alpha = 1.5\)). However, higher watermark strength also introduces perceptible artifacts, leading to a PSNR drop from \(55\,\text{dB}\) to \(42\,\text{dB}\). Bit accuracy remains consistently above \(99\%\) across all settings, confirming the stability of FlowMark’s embedding-decoding pipeline. 

\section{Training Distortions}

Each of the stages in our training use different augmentations for a stabilized training. For each noise type, we list the operation and the range used during training.:

\minisection{Stage 1.}
\begin{itemize}
    \item No distortions; training on clean frames only (\texttt{Identity()}).
\end{itemize}

\minisection{Stage 2}
\paragraph{Valuemetric Transformations.}
\begin{itemize}
    \item Brightness: random in $[0.5,\, 1.5]$ (severity high)
    \item Contrast: random in $[0.5,\, 1.5]$
    \item Hue: random in $[-0.05,\, 0.05]$
    \item Saturation: random in $[0.5,\, 1.5]$
    \item Color jiggle (brightness, contrast, saturation, hue): high severity
    \item Grayscale: applied with probability $0.5$
    \item Gaussian blur: kernel size $k \in \{3, 5, 7\}$, sigma in $(0.1,\, 2.0)$ (high)
    \item Box blur: kernel size $k \in \{3, 5, 7\}$
    \item Median blur: kernel size $3 \times 3$
    \item Gaussian noise: std $\in \{0.02,\, 0.04,\, 0.08\}$ (high)
    \item Motion blur: kernel length $\in \{5, 7, 9\}$, angle/direction by severity
    \item Posterize: severity-dependent bits
    \item RGB shift and sharpness: severity high
\end{itemize}

\paragraph{Image Compression.}
\begin{itemize}
    \item JPEG compression: quality $Q$ in $[40,\, 100]$ (min quality $40$ at high severity)
\end{itemize}

\paragraph{Geometric Transformations.}
\begin{itemize}
    \item Horizontal flip: applied with probability $0.5$
    \item Resized crop: scale $r \in [0.7,\, 1.0]$, aspect ratio in $[3/4,\, 4/3]$
    \item Down–up resize: scale $s \in [0.5,\, 0.85]$ (ramped from gentle to full over training)
\end{itemize}

\minisection{Stage 3}
\paragraph{Valuemetric Transformations.}
\begin{itemize}
    \item Brightness: random in $[0.9,\, 1.1]$
    \item Contrast: random in $[0.9,\, 1.1]$
    \item Hue: random in $[-0.01,\, 0.01]$
    \item Saturation: random in $[0.9,\, 1.1]$
    \item Color jiggle and sharpness: low severity
    \item One optional transform drawn per sample from the above
\end{itemize}

\paragraph{Video Compression.}
\begin{itemize}
    \item VP9: CRF in $[28,\, 40]$
    \item H.264: CRF in $[30,\, 40]$
    \item H.265: CRF in $[32,\, 40]$
    \item AV1: CRF in $[28,\, 40]$
\end{itemize}

\paragraph{Geometric and Temporal.}
\begin{itemize}
    \item Down–up resize: scale $s \in [0.5,\, 0.85]$ (ramped)
    \item Window averaging (temporal): window size in $\{2, 3, 4\}$, blend $\alpha \in [0.25,\, 0.6]$, applied with probability $0.2$ (VP9 branch)
    \item Drop frame: frame replaced by neighbour with probability $0.12$, applied with probability $0.2$ (H.264 branch)
\end{itemize}

\minisection{Combined Transformations (Evaluation Only).}
\begin{itemize}
    \item H.$264$ compression at CRF = $30$
    \item Brightness adjustment with strength $0.5$
    \item Crop removing $50\%$ of the frame area
\end{itemize}

\section{Architecture}

FlowMark consists of three coordinated components: a lightweight mask predictor, an encoder that embeds a binary message into an input image, and a decoder that reconstructs the message from the watermarked content. All modules operate at a spatial resolution of $256\times256$.

\minisection{Mask Predictor.}
The mask predictor is a lightweight convolutional encoder – decoder network that takes a $3$-channel RGB image and outputs both a soft mask and its binarized counterpart. The encoder consists of two convolutional layers with ReLU activations followed by max pooling to capture coarse spatial structure. The decoder upsamples the feature map using bilinear interpolation and refines it through a sequence of $1\times1$ and $3\times3$ convolutional layers with ReLU activations, followed by a final sigmoid layer to produce a soft spatial mask. The final convolution is initialized with a positive bias to encourage stable initial mask activation. During training, a straight-through estimator (STE) converts the soft mask into a binary mask while preserving gradient flow. The predicted mask is used to localize watermark embedding, enabling content-adaptive spatial placement.

\minisection{Encoder.}
The encoder operates on the input image and the bit-vector message. 
The message is first projected through a fully connected layer of size $l \rightarrow L^2$, reshaped into an $L \times L$ grid, and upsampled to the full image resolution using nearest-neighbor interpolation. 
This upsampled grid is passed through a three-layer convolutional stack (Conv--Norm--ReLU) to generate spatial message features, which serve as conditioning signals for embedding.

The input image and message features are concatenated channel-wise and processed by a UNet backbone with four downsampling and upsampling stages, skip connections, and convolutional blocks at each resolution. 
The UNet predicts an intermediate embedding frame, from which a residual is computed. 
This residual is modulated by a JND-based perceptual scaling module and further attenuated using luminance-adaptive dark-region masking to reduce visible artifacts in low-intensity regions. 
The final watermarked image is obtained by adding the modulated residual to the original frame.

To further enhance perceptual realism, we optionally employ GAN-based adversarial supervision following TrustMark~\cite{trustmarkarxiv}. 
A DCGAN-style discriminator with four convolutional layers (base dimension 64) distinguishes original and watermarked frames, encouraging the encoder to generate visually indistinguishable outputs.

\minisection{Decoder.}
The decoder recovers the embedded message from the distorted watermarked frame. 
It first estimates a soft spatial mask using a lightweight U$^2$Net-style subnet composed of convolutional blocks with multi-scale skip connections, producing a single-channel sigmoid mask.

The predicted mask is applied to the input to focus feature extraction on relevant regions. 
The masked image is processed through a shallow convolutional stem (Conv--Norm--ReLU), followed by a UNet backbone symmetric to the encoder architecture. 
The extracted feature map is projected through a small convolutional stack, downsampled to an $L \times L$ grid, flattened, and passed through a fully connected layer to reconstruct the original $l$-bit message.

This dual-branch design allows the decoder to jointly reconstruct the embedding mask and the secret message, improving robustness under compression and geometric distortions.

%
%
\bibliographystyle{splncs04}
\bibliography{main}

@String(CVPR= {IEEE Conf. Comput. Vis. Pattern Recog.})

@String(ICCV= {Int. Conf. Comput. Vis.})

@String(ECCV= {Eur. Conf. Comput. Vis.})

@String(BMVC= {Brit. Mach. Vis. Conf.})

@String(TMM  = {IEEE Trans. Multimedia})

@String(ICASSP=	{ICASSP})

@String(ICIP = {IEEE Int. Conf. Image Process.})

@String(ICLR = {Int. Conf. Learn. Represent.})

@String(CVPR  = {CVPR})

@String(ICCV  = {ICCV})

@String(ECCV  = {ECCV})

@String(BMVC  =	{BMVC})

@String(TMM   =	{IEEE TMM})

@String(ICIP  = {ICIP})

@String(ICLR  = {ICLR})

@article{Hartung1998,
  author    = {F. Hartung and B. Girod},
  title     = {Watermarking of uncompressed and compressed video},
  journal   = {Signal Processing},
  volume    = {66},
  number    = {3},
  pages     = {283--301},
  year      = {1998},
  month     = {May},
}

@article{Coria2008,
  author    = {L. E. Coria and M. R. Pickering and P. Nasiopoulos and R. K. Ward},
  title     = {A video watermarking scheme based on the dual-tree complex wavelet transform},
  journal   = {IEEE Transactions on Information Forensics and Security},
  volume    = {3},
  number    = {3},
  pages     = {466--474},
  year      = {2008},
}

@article{RivaGAN2019,
  author    = {H. Zhang and J. Dong and P. Forgione and J. Collomosse},
  title     = {RivaGAN: Robust Video Watermarking with Attention},
  journal   = {arXiv preprint arXiv:1909.01230},
  year      = {2019},
}

@article{DVMark2023,
  author    = {X. Luo and Y. Li and H. Chang and C. Liu and P. Milanfar and F. Yang},
  title     = {DVMark: A Deep Multiscale Framework for Video Watermarking},
  journal   = {IEEE Transactions on Image Processing},
  year      = {2023},
}

@inproceedings{VideoSeal2024,
  author    = {J. Chang and S. Choi and P. Milanfar},
  title     = {VideoSeal: A Learned Model for Robust Video Watermarking},
  booktitle = {Proceedings of the IEEE/CVF Conference on Computer Vision and Pattern Recognition (CVPR)},
  year      = {2024},
}

@article{Alattar2003,
  author    = {A. M. Alattar and E. T. Lin and M. U. Celik},
  title     = {Digital Watermarking of Low Bit-Rate Advanced Simple Profile MPEG-4 Compressed Video},
  journal   = {IEEE Transactions on Circuits and Systems for Video Technology},
  volume    = {13},
  number    = {8},
  pages     = {787--800},
  year      = {2003},
  month     = {Aug},
}

@inproceedings{Yong2008,
  author    = {M. Yong and T. Yu-Min and Q. Yun-Hui},
  title     = {Adaptive Video Watermarking Algorithm Based on MPEG-4 Streams},
  booktitle = {Proceedings of the International Conference on Control, Automation, Robotics and Vision},
  pages     = {1084--1088},
  year      = {2008},
}

@article{Zhang2007,
  author    = {J. Zhang and A. T. S. Ho and G. Qiu and P. Marziliano},
  title     = {Robust Video Watermarking of H.264/AVC},
  journal   = {IEEE Transactions on Circuits and Systems II: Express Briefs},
  volume    = {54},
  number    = {2},
  pages     = {205--209},
  year      = {2007},
}

@article{Guo2010,
  author    = {J. Guo and Y. Pan},
  title     = {Motion Vector-Based Watermarking for H.264 Video Stream with Drift Compensation},
  journal   = {IEEE Transactions on Circuits and Systems for Video Technology},
  volume    = {20},
  number    = {8},
  pages     = {1164--1170},
  year      = {2010},
}

@article{Tew2016,
  author    = {Y. Tew and K. Wong},
  title     = {An Efficient Video Watermarking Technique for HEVC Video Authentication},
  journal   = {IEEE Transactions on Circuits and Systems for Video Technology},
  volume    = {26},
  number    = {1},
  pages     = {65--76},
  year      = {2016},
}

@inproceedings{Dutta2018,
  author    = {A. Dutta and S. Gupta},
  title     = {A Blind and Robust Video Watermarking Technique for H.265/HEVC Encoded Video},
  booktitle = {Proceedings of the IEEE International Conference on Computing, Communication and Automation},
  pages     = {1189--1194},
  year      = {2018},
}

@misc{synthid,
  title     = {Identifying {A}{I}-generated images with {SynthID}},
  author    = {{Google DeepMind}},
  year      = {2023},
  howpublished = {\url{https://deepmind.google/discover/blog/identifying-ai-generated-images-with-synthid/}}
}

@article{Collomosse2024,
  author    = {J. Collomosse and A. Parsons},
  title     = {{To Authenticity, and Beyond! Building Safe and Fair Generative AI upon the Three Pillars of Provenance}},
  journal   = {IEEE Computer Graphics and Applications (IEEE CG\&A)},
  year      = {2024}
}

@techreport{c2pa,
  author      = "{Coalition for Content Provenance and Authenticity}",
  title       = "Technical Specification 2.2",
  institution = "C2PA",
  year        = "2025",
  url = "https://c2pa.org/specifications/specifications/2.2/index.html"
}

@inproceedings{hu2025mask,
  title={Mask Image Watermarking},
  author={Runyi Hu and Jie Zhang and Shiqian Zhao and Nils Lukas and Jiwei Li and Qing Guo and Han Qiu and Tianwei Zhang},
  booktitle={The Thirty-ninth Annual Conference on Neural Information Processing Systems},
  year={2025}
}

@article{wu2017enhanced,
  title={Enhanced just noticeable difference model for images with pattern complexity},
  author={Wu, Jinjian and Li, Leida and Dong, Weisheng and Shi, Guangming and Lin, Weisi and Kuo, C-C Jay},
  journal={IEEE Transactions on Image Processing},
  volume={26},
  number={6},
  pages={2682--2693},
  year={2017},
  publisher={IEEE}
}

@inproceedings{bui2025trustmark,
  title={TrustMark: Robust Watermarking and Watermark Removal for Arbitrary Resolution Images},
  author={Bui, Tu and Agarwal, Shruti and Collomosse, John},
  booktitle={Proceedings of the IEEE/CVF International Conference on Computer Vision},
  pages={18629--18639},
  year={2025}
}

@inproceedings{bui2023rosteals,
  title={{RoSteALS}: Robust Steganography using Autoencoder Latent Space},
  author={Bui, Tu and Agarwal, Shruti and Yu, Ning and Collomosse, John},
  booktitle={CVPR},
  year={2023}
}

@inproceedings{balan2023ekila,
  title={{EKILA}: Synthetic Media Provenance and Attribution for Generative Art},
  author={Balan, Kar and Agarwal, Shruti and Jenni, Simon and Parsons, Andy and Gilbert, Andrew and Collomosse, John},
  booktitle={CVPR},
  year={2023}
}

@inproceedings{asnani2025custommark,
  title={CustomMark: Customization of Diffusion Models for Proactive Attribution},
  author={V. Asnani and J. Collomosse and X. Liu and S. Agarwal},
  booktitle={Proc. ICCV Workshop on Authenticity \& Provenance in the age of Generative AI (APAI)},
  year={2025}
}

@inproceedings{nguyen2021,
  title={{OSCAR-Net}: Object-centric Scene Graph Attention for Image Attribution},
  author={Eric Nguyen and Tu Bui and Vishy Swaminathan and John Collomosse},
  booktitle={ICCV},
  year={2021}
}

@inproceedings{asnani2022proactive,
  title={Proactive image manipulation detection},
  author={Asnani, Vishal and Yin, Xi and Hassner, Tal and Liu, Sijia and Liu, Xiaoming},
  booktitle={CVPR},
  year={2022}
}

@inproceedings{fernandez2023stable,
  title={The stable signature: Rooting watermarks in latent diffusion models},
  author={Fernandez, Pierre and Couairon, Guillaume and J{\'e}gou, Herv{\'e} and Douze, Matthijs and Furon, Teddy},
  booktitle={ICCV},
  year={2023}
}

@inproceedings{asnani2024promark,
  title={ProMark: Proactive Diffusion Watermarking for Causal Attribution},
  author={Asnani, Vishal and Collomosse, John and Bui, Tu and Liu, Xiaoming and Agarwal, Shruti},
  booktitle={Proceedings of the IEEE/CVF Conference on Computer Vision and Pattern Recognition},
  pages={10802--10811},
  year={2024}
}

@inproceedings{Petrov-NeurIPS-2025,
  author = {A. Petrov and S. Agarwal and P. Torr and A. Bibi and J. Collomosse},
  title = {{On the Coexistence and Ensembling of Watermarks}},
  year = {2025},
  booktitle = {Intl. Conf. Neural Information Processing Systems (NeurIPS), 2025},
  url = {https://collomosse.com/assets/pubs/Petrov-NeurIPS-2025.pdf}
}

@article{videosealarxiv,
title={{VideoSeal}: Open and Efficient Video Watermarking},
author={Pierre Fernandez and Hady Elsahar and I. Zeki Yalniz and Alexandre Mourachko},
journal = {ArXiv e-prints},
archivePrefix = "arXiv",
eprint = {2412.09492},
year = 2024,
month = dec
}

@article{Bharati2021,
  title={Transformation-Aware Embeddings for Image Provenance},
  author={Bharati, A. and Moreira, D. and Flynn, P. and de Rezende Rocha, A. and Bowyer, K. and Scheirer, W.},
  journal={IEEE Trans. Info. Forensics and Sec.},
  year={2021},
  pages={2493--2507},
  volume={16},
  publisher={IEEE}
}

@article{Zhang2020manip,
  title={Discovering Image Manipulation History by Pairwise Relation and Forensics Tools},
  author={Zhang, X. and Sun, Z. H. and Karaman, S. and Chang, S.F.},
  journal={IEEE J. Selected Topics in Signal Processing.},
  year={2020},
  pages={1012--1023},
  volume={14},
  number={5},
  publisher={IEEE}
}

@article{devi2009,
  title={A Fragile Watermarking scheme for Image Authentication with Tamper Localization Using Integer Wavelet transform},
  author={P. Devi and M. Venkatesan and K. Duraiswamy},
  journal={J. Computer Science},
  volume={5},
  number={11},
  pages={831--837},
  year={2019}
}

@article{baba2009,
  title={Watermarking Scheme for copyright protection of digital images},
  author={S. Baba and L. Krekor and T. Arif and Z.  Shaaban},
  journal={IJCSNS},
  volume={9},
  number={4},
  year={2019}
}

@inproceedings{weng2019high,
  title={High-capacity convolutional video steganography with temporal residual modeling},
  author={Weng, Xinyu and Li, Yongzhi and Chi, Lu and Mu, Yadong},
  booktitle={Proc. ICMR},
  pages={87--95},
  year={2019}
}

@inproceedings{wolfgang1996watermark,
  title={A watermark for digital images},
  author={Wolfgang, Raymond B and Delp, Edward J},
  booktitle={Proc. ICIP},
  volume={3},
  pages={219--222},
  year={1996},
  organization={IEEE}
}

@article{taha2022high,
  title={High payload image steganography scheme with minimum distortion based on distinction grade value method},
  author={Taha, Mustafa Sabah and Rahem, Mohd Shafry Mohd and Hashim, Mohammed Mahdi and Khalid, Hiyam N},
  journal={Multimedia Tools and Applications},
  volume={81},
  number={18},
  pages={25913--25946},
  year={2022},
  publisher={Springer}
}

@inproceedings{ghazanfari2011lsb++,
  title={LSB++: An improvement to LSB+ steganography},
  author={Ghazanfari, Kazem and Ghaemmaghami, Shahrokh and Khosravi, Saeed R},
  booktitle={TENCON 2011-2011 IEEE Region 10 Conference},
  pages={364--368},
  year={2011},
  organization={IEEE}
}

@inproceedings{navas2008dwt,
  title={{DWT}-{DCT}-{SVD} based watermarking},
  author={Navas, KA and Ajay, Mathews Cheriyan and Lekshmi, M and Archana, Tampy S and Sasikumar, M},
  booktitle={COMSWARE'08},
  pages={271--274},
  year={2008},
  organization={IEEE}
}

@inproceedings{fernandez2022watermarking,
  title={Watermarking images in self-supervised latent spaces},
  author={Fernandez, Pierre and Sablayrolles, Alexandre and Furon, Teddy and J{\'e}gou, Herv{\'e} and Douze, Matthijs},
  booktitle={Proc. ICASSP},
  pages={3054--3058},
  year={2022},
  organization={IEEE}
}

@inproceedings{zhu2018hidden,
  title={Hidden: Hiding data with deep networks},
  author={Zhu, Jiren and Kaplan, Russell and Johnson, Justin and Fei-Fei, Li},
  booktitle={Proc. ECCV},
  pages={657--672},
  year={2018}
}

@inproceedings{zhang2025omniguard,
  title={OmniGuard: Hybrid Manipulation Localization via Augmented Versatile Deep Image Watermarking},
  author={Xuanyu Zhang and Zecheng Tang and Zhipei Xu and Runyi Li and Youmin Xu and Bin Chen and Feng Gao and Jian Zhang},
  booktitle={Proc. CVPR},
  year={2025}
}

@inproceedings{archangel1,
  author = {T. Bui and D. Cooper and J. Collomosse and M. Bell and A. Green and J. Sheridan and J. Higgins and A. Das and J. Keller and O. Thereaux and A. Brown},
  title = {{ARCHANGEL: Tamper-proofing Video Archives using Temporal Content Hashes on the Blockchain}},
  year = {2019},
  booktitle = {CVPR Workshops (Computer Vision, AI and Blockchain), 2019},
  url = {https://collomosse.com/assets/pubs/Bui-CVPRWS-2019.pdf}
}

@article{archangel2,
  author = {T. Bui and D. Cooper and J. Collomosse and M. Bell and A. Green and J. Sheridan and J. Higgins and A. Das and J. Keller and O. Thereaux},
  title = {{Tamper-proofing Video with Hierarchical Attention Autoencoder Hashing on Blockchain}},
  year = {2020},
  journal = {IEEE Transactions on Multimedia (TMM), 2020},
  url = {https://collomosse.com/assets/pubs/Bui-TMM-2020.pdf}
}

@inproceedings{sander2025wam,
  title={Watermark Anything With Localized Messages},
  author={Tom Sander and Pierre Fernandez and Alain Oliviero Durmus and Teddy Furon and Matthijs Douze},
  booktitle={Proc. ICLR},
  year={2025}
}

@article{trustmarkarxiv,
title={Trustmark: Universal Watermarking for Arbitrary Resolution Images},
author={Bui, Tu and Agarwal, Shruti and Collomosse, John},
journal = {ArXiv e-prints},
archivePrefix = "arXiv",
eprint = {2311.18297},
year = 2023,
month = nov
}

@inproceedings{decorait,
  author = {K. Balan and A. Black and S. Jenni and A. Gilbert and A. Parsons and J. Collomosse},
  title = {{DECORAIT - DECentralized Opt-in/out Registry for AI Training}},
  year = {2023},
  booktitle = {Conference on Visual Media Production (CVMP), 2023},
  url = {https://collomosse.com/assets/pubs/Balan-CVMP-2023.pdf}
}

@misc{accct,
  author = {J. Bennett and J. Collomosse and R. Gregory-Clarke and J. Jones and L. Love and M. Lycett and W. Saunders},
  title = {{Time to ACCCT: Providing Creative Industries and AI Developers with a Copyright Framework of Access, Control, Consent, Compensation and Transparency}},
  year = {2025},
  howpublished = {CoSTAR/DECaDE/Sheridans Technical Report},
  url = {https://collomosse.com/assets/pubs/Bennett-ACCCT-2025.pdf}
}

@inproceedings{contentarcs,
  author = {K. Balan and A. Gilbert and J. Collomosse},
  title = {{Content ARCs: Decentralized Content Rights in the Age of Generative AI}},
  year = {2025},
  booktitle = {International Conference on AI and the Digital Economy (CADE), 2025},
  url = {https://collomosse.com/assets/pubs/Balan-CADE-2025.pdf}
}

@article{wan2022comprehensive,
  title={A comprehensive survey on robust image watermarking},
  author={Wan, Wenbo and Wang, Jun and Zhang, Yunming and Li, Jing and Yu, Hui and Sun, Jiande},
  journal={Neurocomputing},
  year={2022},
  publisher={Elsevier}
}

@inproceedings{tancik2020stegastamp,
  title={Stegastamp: Invisible hyperlinks in physical photographs},
  author={Tancik, Matthew and Mildenhall, Ben and Ng, Ren},
  booktitle={Proc. CVPR},
  pages={2117--2126},
  year={2020}
}

@inproceedings{xu2025invismark,
  title={InvisMark: Invisible and Robust Watermarking for AI-generated Image Provenance},
  author={Xu, Rui and Hu, Mengya and Lei, Deren and Li, Yaxi and Lowe, David and Gorevski, Alex and Wang, Mingyu and Ching, Emily and Deng, Alex and others},
  booktitle={Proc. Winter Conf. on Appl. of Computer Visoin (WACV)},
  year={2025}
}

@inproceedings{wen2023treerings,
  title     = {Tree‑Rings Watermarks: Invisible Fingerprints for Diffusion Images},
  author    = {Yuxin Wen and John Kirchenbauer and Jonas Geiping and Tom Goldstein},
  booktitle = {Advances in Neural Information Processing Systems (NeurIPS)},
  volume    = {36},
  year      = {2023},
}

@article{hu2025maskimagewatermarking,
  title        = {Mask Image Watermarking},
  author       = {Runyi Hu and Jie Zhang and Shiqian Zhao and Nils Lukas and Jiwei Li and Qing Guo and Han Qiu and Tianwei Zhang},
  year         = {2025},
  journal={arXiv preprint arXiv:2405.11135},
}

@misc{jpegtrust2025,
  title        = {An Update on {J}{P}{E}{G} Trust},
  author       = {{P. Rixhon}},
  year         = {2025},
  month         = {January},
  howpublished = {\url{https://cawg.io/meeting-notes/_attachments/2025-01-21/jpeg-trust-presentation.pdf}},
}

@inproceedings{mishra2019vstegnet,
  title={VStegNET: Video Steganography Network using Spatio-Temporal features and Micro-Bottleneck.},
  author={Mishra, Aayush and Kumar, Suraj and Nigam, Aditya and Islam, Saiful},
  booktitle={BMVC},
  volume={274},
  year={2019}
}

@article{shen2023vhnet,
  title={VHNet: A video hiding network with robustness to video coding},
  author={Shen, Xiaofeng and Yao, Heng and Tan, Shunquan and Qin, Chuan},
  journal={Journal of Information Security and Applications},
  volume={75},
  pages={103515},
  year={2023},
  publisher={Elsevier}
}

@inproceedings{zhang2023novel,
  title={A novel deep video watermarking framework with enhanced robustness to H. 264/AVC compression},
  author={Zhang, Yulin and Ni, Jiangqun and Su, Wenkang and Liao, Xin},
  booktitle={Proceedings of the 31st ACM International Conference on Multimedia},
  pages={8095--8104},
  year={2023}
}

@article{jia2022rivie,
  title={RIVIE: Robust inherent video information embedding},
  author={Jia, Jun and Gao, Zhongpai and Zhu, Dandan and Min, Xiongkuo and Hu, Menghan and Zhai, Guangtao},
  journal={IEEE Transactions on Multimedia},
  volume={25},
  pages={7364--7377},
  year={2022},
  publisher={IEEE}
}

@inproceedings{zhang2024v2a,
  title={V2a-mark: Versatile deep visual-audio watermarking for manipulation localization and copyright protection},
  author={Zhang, Xuanyu and Xu, Youmin and Li, Runyi and Yu, Jiwen and Li, Weiqi and Xu, Zhipei and Zhang, Jian},
  booktitle={Proceedings of the 32nd ACM International Conference on Multimedia},
  pages={9818--9827},
  year={2024}
}

@inproceedings{ye2023itov,
  title={ItoV: efficiently adapting deep learning-based image watermarking to video watermarking},
  author={Ye, Guanhui and Gao, Jiashi and Wang, Yuchen and Song, Liyan and Wei, Xuetao},
  booktitle={2023 International Conference on Culture-Oriented Science and Technology (CoST)},
  pages={192--197},
  year={2023},
  organization={IEEE}
}

@article{fernandez2024video,
  title={Video seal: Open and efficient video watermarking},
  author={Fernandez, Pierre and Elsahar, Hady and Yalniz, I Zeki and Mourachko, Alexandre},
  journal={arXiv preprint arXiv:2412.09492},
  year={2024}
}

@inproceedings{mbrs,
author = {Jia, Zhaoyang and Fang, Han and Zhang, Weiming},
title = {MBRS: Enhancing Robustness of DNN-based Watermarking by Mini-Batch of Real and Simulated JPEG Compression},
year = {2021},
booktitle = {Proceedings of the 29th ACM International Conference on Multimedia},
}

@inproceedings{ma2022towards,
  title={Towards blind watermarking: Combining invertible and non-invertible mechanisms},
  author={Ma, Rui and Guo, Mengxi and Hou, Yi and Yang, Fan and Li, Yuan and Jia, Huizhu and Xie, Xiaodong},
  booktitle={Proceedings of the 30th ACM International Conference on Multimedia},
  pages={1532--1542},
  year={2022}
}

@misc{netflix2019video,
  title={Video multi-method assessment fusion},
  author={Netflix, VMAF},
  year={2019}
}

@inproceedings{kirillov2023segment,
  title={Segment anything},
  author={Kirillov, Alexander and Mintun, Eric and Ravi, Nikhila and Mao, Hanzi and Rolland, Chloe and Gustafson, Laura and Xiao, Tete and Whitehead, Spencer and Berg, Alexander C and Lo, Wan-Yen and others},
  booktitle={Proceedings of the IEEE/CVF international conference on computer vision},
  pages={4015--4026},
  year={2023}
}
\end{document}